\documentclass[10pt,twocolumn,letterpaper]{article}

\usepackage{cvpr}              

\definecolor{cvprblue}{rgb}{0.21,0.49,0.74}
\usepackage[pagebackref,breaklinks,colorlinks,allcolors=cvprblue]{hyperref}
\usepackage{multirow}
\usepackage{bbding}
\usepackage{graphicx}
\usepackage{booktabs}
\usepackage{xcolor}
\usepackage{colortbl}

\usepackage{amsmath}
\usepackage{algorithm}
\usepackage{algpseudocode}

\def\paperID{3729} 
\def\confName{CVPR}
\def\confYear{2025}

\title{
Segment Any-Quality Images with Generative Latent Space Enhancement
}
\author{
        Guangqian Guo$^{1,2}$ \enspace 
        Yong Guo$^2$ \enspace 
        Xuehui Yu$^3$ \enspace 
        Wenbo Li$^4$  \enspace 
        Yaoxing Wang$^1$ \enspace
        Shan Gao$^{1}$  \enspace  \\
        \textsuperscript{1}Northwestern Polytechnical University \enspace
        \textsuperscript{2}Huawei \enspace \\
        \textsuperscript{3}University of Chinese Academy of Sciences \enspace 
        \textsuperscript{4} Huawei Noah's Ark Lab \\ 
        \{guogq21, wangyx24\}@mail.nwpu.edu.cn \enspace  
        \{guoyongcs, fenglinglwb\}@gmail.com \\
        yuxuehui17@mails.ucas.ac.cn \enspace
        gaoshan@nwpu.edu.cn
}

\begin{document}
\renewcommand{\thefootnote}{\fnsymbol{footnote}}

\maketitle
\footnotetext[1] {This work was done during Guangqian Guo's internship at Huawei, supervised by Yong Guo. The first two authors contribute equally.}  
\footnotetext[2]{Corresponding author: Shan Gao}

\begin{abstract}
Despite their success, Segment Anything Models (SAMs) experience significant performance drops on severely degraded, low-quality images, limiting their effectiveness in real-world scenarios. To address this, we propose \textbf{GleSAM}, which utilizes \underline{G}enerative \underline{L}atent space \underline{E}nhancement to boost robustness on low-quality images, thus enabling generalization across various image qualities.  Specifically, we adapt the concept of latent diffusion to SAM-based segmentation frameworks and perform the generative diffusion process in the latent space of SAM to reconstruct high-quality representation, thereby improving segmentation. Additionally, we introduce two techniques to improve compatibility between the pre-trained diffusion model and the segmentation framework. Our method can be applied to pre-trained SAM and SAM2 with only minimal additional learnable parameters, allowing for efficient optimization.
We also construct the LQSeg dataset with a greater diversity of degradation types and levels for training and evaluating the model. Extensive experiments demonstrate that GleSAM significantly improves segmentation robustness on complex degradations while maintaining generalization to clear images. Furthermore, GleSAM also performs well on unseen degradations, underscoring the versatility of our approach and dataset. Code and dataset are released at \href{https://github.com/guangqian-guo/GleSAM}{\textcolor{red}{\textit{This Page}}}.
\end{abstract}

\section{Introduction}
\label{sec:intro}

\begin{figure}[t]
  \centering
   \includegraphics[width=1 \linewidth]{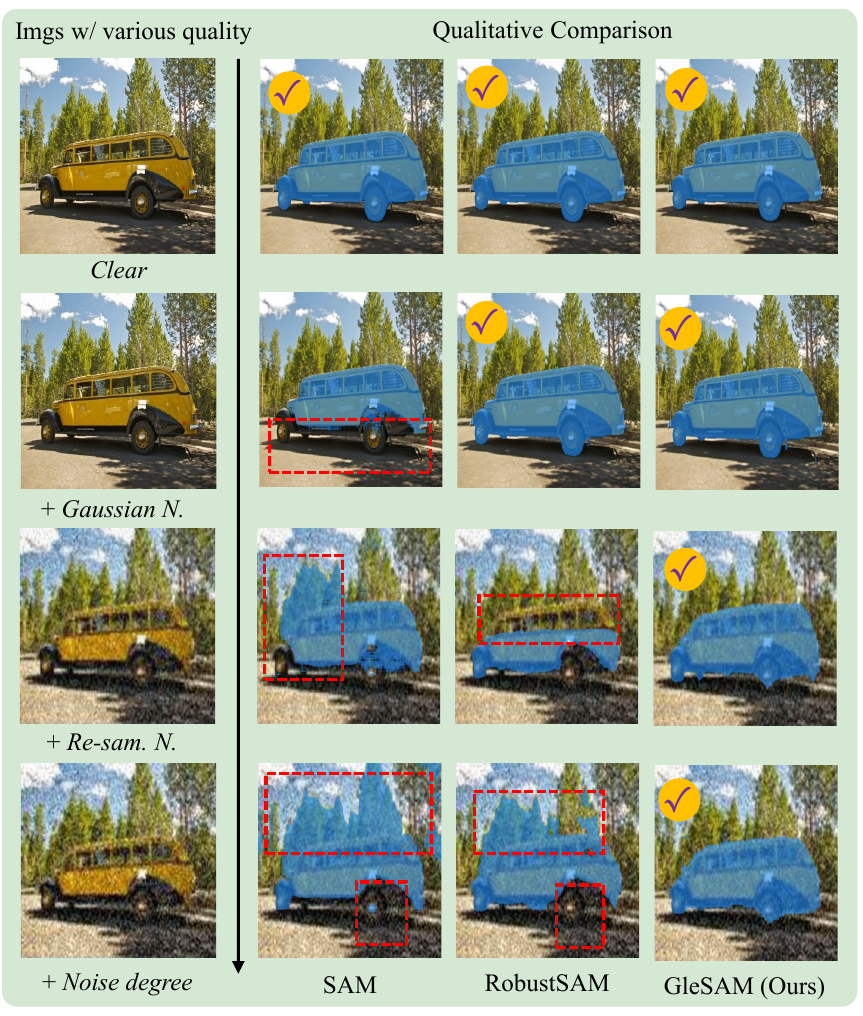}
   \caption{The comparison of qualitative results on low-quality images with varying degradation levels from an unseen dataset. To generate images with different degradation levels, we progressively added Gaussian Noise, Re-sampling Noise, and more severe Gaussian noise to an image. Results indicate that the baseline SAM \cite{sam} shows limited robustness to degradation. Although RobustSAM \cite{robustsam} retains some resilience against simpler degradations, it struggles with more complex and unfamiliar degradations. In contrast, our method consistently demonstrates strong robustness across images of varying quality.}
   \label{fig:comparison}
   \vspace{-10pt}
\end{figure}

Accurate object detection and segmentation~\cite{maskrcnn, panoptic, p2p, videoseg, pcod, hybridnet, hanet, effective-rotate} in diverse scenarios is a fundamental task for various high-level visual applications, such as robotics and autonomous driving. The recently developed Segment Anything Models (SAMs), including SAM \cite{sam} and SAM2 \cite{sam2}, serving as a foundational model, have gained significant influence within the community ~\cite{p2p, rsprompter, sam-cod, sam-rs3, sam-med, sam-med2} due to their outstanding zero-shot segmentation abilities. 

\begin{figure}[t]
  \centering
   \includegraphics[width=1\linewidth]{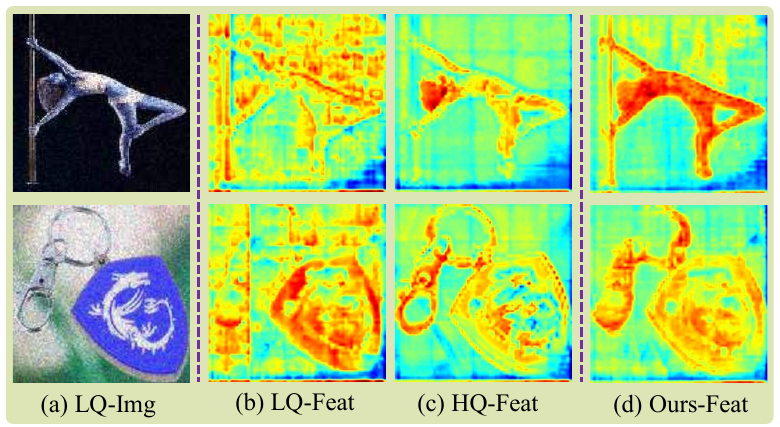}
   \caption{
   The visualization of latent features: (a) low-quality (LQ) images, (b) the SAM's latent features extracted from LQ images, which contain excessive noise and compromise the original representations, (c) the high-quality (HQ) features of the corresponding clear images, which are more salient than LQ ones, and (d) enhanced representation by our GleSAM.
   }
   \label{fig:lq-feat}
\end{figure}

{Despite their success, SAMs perform poorly on common low-quality images, such as those degraded by noise, blur, and compression artifacts \cite{sam-robustness1, sam-robustness2, sam-robustness3, robustness4}, which are often encountered in real-world scenarios \cite{image-noise, image-noise2, image-noise3}.                        
Previous methods \cite{robustsam, lwfa} have employed distillation-based consistent learning to enhance degradation-robust features. 
Nonetheless, they still face challenges in handling severely degraded low-quality images, as illustrated in Fig.~\ref{fig:comparison}. As degradations become more complex (\eg combining various types of degradation or increasing the level of degradation), the existing SAMs \cite{sam, robustsam} struggle to accurately segment edges and complete target areas, leading to incorrect segmentation.
We analyze that it is caused by the limited feature representation for degraded images.
The visualizations in Fig.~\ref{fig:lq-feat} reveal that SAM's latent features from severely degraded images contain excessive noise, compromising the original representations and subsequently impacting the predictions of the decoder. Furthermore, the large gap between low-quality and high-quality features complicates consistency learning \cite{takd} in previous works \cite{robustsam}, hindering performance improvement. Thus, achieving high-quality latent feature representations and robust segmentation across varying image quality, especially for degraded images, remains challenging.
}

The recently developed generative Diffusion Models (DM) \cite{ddpm, ddim}, especially the large-scale pre-trained Latent Diffusion Models (LDM) \cite{ldm} have demonstrated powerful content generation capabilities. Having been trained on internet-scale data \cite{laion}, LDM that proceed diffusion and denoising in latent space, possess powerful representation prior, which can be well explored to enhance the latent representation of segmentation models.
This inspires us to take full advantage of the generative ability of pre-trained diffusion models and incorporate it into the latent space of SAMs to enhance low-quality features, thus promoting accurate segmentation in low-quality images.

To this end, we propose \textbf{\textit{GleSAM}}, which reconstructs high-quality features (Fig. \ref{fig:lq-feat} (d)) in SAM's latent space through generative diffusion, enabling accurate segmentation across any-quality images. Starting with low-quality features, high-quality representations are generated through single-step denoising. To integrate LDM generative knowledge, we incorporate a pre-trained U-Net from LDM with learnable LoRA layers \cite{lora} to align with segmentation-specific features.
Furthermore, to improve compatibility between the pre-trained diffusion model and the segmentation framework, we introduce two effective techniques: \textbf{\textit{Feature Distribution Alignment}} (FDA) and \textbf{\textit{Channel Replicate and Expansion}} (CRE). These techniques bridge feature distribution and structural alignment gaps between models. Built upon SAM and SAM2, GleSAM leverages the generalization of pre-trained segmentation and diffusion models, with a few learnable parameters added, and can be efficiently trained within 30 hours on four GPUs.

In terms of data, we constructed \textbf{\textit{LQSeg}} based on existing datasets \cite{thin, lvis, coco, msra, ecssd} to train and assess segmentation models on low-quality images.  
LQSeg incorporates a greater diversity of degradation types than previous methods \cite{robustsam}, combining basic degradation models (e.g., noise and blur) to simulate complex and real-world noise \cite{realesrgan, bsrgan}. We also introduce three degradation levels for a more comprehensive evaluation. We hope LQSeg will inspire the development of more robust segmentation models and contribute to future research.
Overall, our contributions are summarized as: 
\begin{itemize}
    \item We propose GleSAM, a SAM-based framework incorporating generative latent space enhancement, to generalize across images of any quality. GleSAM exhibits significantly improved robustness, particularly for low-quality images with varying degradation levels. 
    \item Two effective techniques: FDA and CRE, are introduced to bridge feature distribution and structural gaps between the pre-trained latent diffusion model and SAM. 
    \item  We also construct the LQSeg dataset which includes a wide range of degradation types and levels, to effectively train and evaluate the model. 
    \item Extensive experiments show that our method performs excellently on low-quality images with varying degrees of degradation while maintaining generalization to clear images. Additionally, our method achieves strong performance on unseen degradations, highlighting the adaptability of both our framework and dataset.
\end{itemize}

\section{Related Work}
\label{sec:related}

\subsection{Segmentation on Low-Quality Images}
Executing robust segmentation across various scenarios is a critical issue.  Numerous studies \cite{benchmark-robustness, improve-robust, sam-robustness1, robustsam} have highlighted significant performance degradation in conventional segmentation models and foundational SAMs when confronted with low-quality images with degradation. 
Many related studies \cite{improve-robust, benchmark-robustness, lwfa, fifo, dense-gram} have been proposed to enhance the robustness of segmentation models against low-quality data. 
These methods primarily consider a single type of degradation. 
Recently, RobustSAM \cite{robustsam} is introduced to enhance the robustness of the SAM against multiple image degradations through anti-degradation feature learning.
However, its performance also struggles when dealing with complex degradations.
The real-world image noise is often too complex to be modeled by a single degradation \cite{image-noise, image-noise2, image-noise3}. Therefore, robustly segmenting images of any quality remains challenging.

\subsection{Diffusion Models for Perception Tasks}
Recently, diffusion models \cite{ddpm, ldm, ddim, dmd, osdiff-vsd} have garnered significant attention in research, due to their powerful generation capabilities. Numerous studies \cite{segdiff, vpd, ddp, ldznet, diffusiondet, segrefiner, lessdm, diffbir, osediff} explore how to extend their applications to a broader range of tasks, such as detection, segmentation, and image reconstruction, \etc. 
For diffusion-based perception tasks, one category of methods \cite{ segdiff,diff-generalist, medsegdiff, medsegdiffv2} reformulate the perception tasks as progressive denoising from random noise. such as DiffusionDet \cite{diffusiondet} and DiffusionInst \cite{diffusiondet}. 
Another route employs the denoising autoencoder pre-trained on the text-to-image generation as a backbone for downstream perception tasks \cite{ddp, vpd, tadp, ldznet, ptdiffseg}. 
For example, VPD \cite{vpd}  passes the image through a pre-trained diffusion model and extracts intermediate features for task prediction. 
Diverging from these existing works, we preserve the original segmentation structure and fine-tune a generative diffusion to enhance the segmentation model’s latent representations for accurate segmentation of any quality images.

\subsection{Segment Anything Model and Variants}
Segment Anything Models (SAMs) ~\cite{sam, sam2} have gained significant influence within the community due to their outstanding zero-shot segmentation capabilities. SAM \cite{sam} can interactively segment any object in an image using visual prompts such as points and bounding boxes. 
Most recently, the updated SAM2 \cite{sam2} has been released, showing improved segmentation accuracy and inference efficiency. 
Their generalization abilities have led to breakthroughs and new paradigms in various downstream tasks \cite{samhq, asam, samrs, samantic-aware-sam, sam-med, medsam2, 3dsam2, gaussian-grouping}. 
Although SAM is powerful, its performance decreases when facing complex scenarios, such as degraded images \cite{sam-robustness1, sam-robustness2, sam-robustness3} and adverse weather conditions \cite{robustness4}, which significantly hinders the real-world applications of SAM.
Enhancing SAM's capability in such challenging scenarios is a worthwhile research topic.

\begin{figure*}[t]
  \centering
   \includegraphics[width=1.\linewidth]{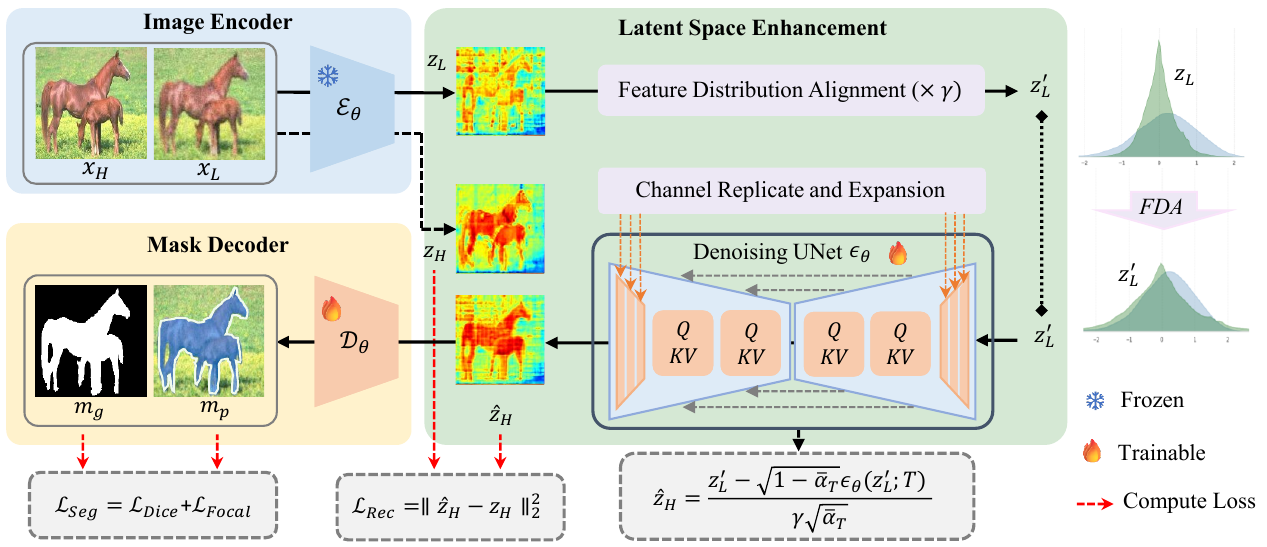}
   \caption{
   Given an input image, GleSAM performs accurate segmentation through image encoding, generative latent space enhancement, and mask decoding. During training, HQ-LQ image pairs are fed into the frozen image encoder to extract the corresponding HQ and LQ latent features. We then reconstruct high-quality representations in the SAM's latent space by efficiently fine-tuning a generative denoising U-Net with LoRA. Subsequently, the decoder is fine-tuned with segmentation loss to align the enhanced latent representations. Built upon SAMs, GleSAM inherits prompt-based segmentation and performs well on images of any quality.
   }
   \vspace{-7pt}
   \label{fig:pipeline}
\end{figure*}

\section{Generative Latent Space Enhancement for Any-Quality Image Segmentation }  
\label{sec:method}

In the following, we explore how to improve SAM's robustness for low-quality images through generative latent space enhancement, thus enabling it to generalize across varying image qualities. The overall framework of the proposed \textbf{\textit{GleSAM}} is shown in Fig. \ref{fig:pipeline}.
To begin, in Sec. \ref{sec: fdm}, we propose incorporating diffusion models' generative capabilities into SAM's latent space to effectively and efficiently enhance low-quality feature representations. Next, to improve the compatibility of feature distribution and architecture between the pre-trained diffusion model and SAM,  we introduce two techniques: \textbf{\textit{Feature Distribution Alignment}} and \textbf{\textit{Channel Replicate and Expansion}}, which are detailed in Sec. \ref{sec: faw} and Sec. \ref{sec: REI}, respectively. Finally, the overall training method is outlined in Sec. \ref{sec: ft}.
 
\subsection{Latent Denoising Diffusion in Segmentation}
\label{sec: fdm}    
Recall that diffusion models \cite{ddpm, ddim, ldm} are a class of probabilistic generative models that progressively add noise to the latent space, and then they learn to reverse this process by predicting and removing the noise. Formally, in LDMs, the forward noise process iteratively adds Gaussian noise with variance $\beta_t \in (0,\mathrm{I})$ to the variable $z$. The sample at each time point is defined as: 
\begin{equation}
    z_t =\sqrt{\overline{\alpha}_{t}}z + \sqrt{1-\overline{\alpha}_t}\epsilon,
    \label{eq:1}
\end{equation}
where $\alpha_t = 1-\beta_t$, $\overline{\alpha}_{t}=\prod_{s=1}^{t}\alpha_s$, and $\epsilon \in \mathcal{N}(0, \mathrm{I})$. 
While the inverse diffusion process is modeled by applying a neural network $\epsilon_{\theta}(z_t, t)$ to predict the noise $\hat{\epsilon}$ and recover the original input $\hat{z}$. 
LDMs model the above process in a latent space using a pre-trained Variational AutoEncoder (VAE) and then up-sample the latent output to the original resolution using the VAE decoder, enabling more efficient computations in the training and inference phases.

A similar idea motivates us to introduce the generative latent space denoising process into the SAMs' framework to reconstruct low-quality segmentation features.  
Let's denote $\mathcal{E}_{\theta}$ and $\mathcal{D}_{\theta}$ the segmentation encoder and decoder of SAMs, respectively. 
As shown in Fig. \ref{fig:pipeline}, the input LQ image is first compressed by $\mathcal{E}_{\theta}$ and generates LQ feature $z_L$. 
We consider $z_L$ to be a noisy version of $z_H$, containing sufficient information to reconstruct a high-quality feature. Instead of the complex multi-step denoising from random noise, we start directly from $z_L$ and forward with a single denoising step. Specifically, based on Eq. \ref{eq:1}, the clean latent variable $z$ can be directly predicted from the model's predicted noise $\hat{\epsilon}$, as: 
\begin{equation}
    \hat{z} = \frac{z_t-\sqrt{1-\overline{\alpha}_t}\hat{\epsilon}}{\sqrt{\overline{\alpha}_t}},
\end{equation}
where $\hat{\epsilon}$ is the prediction of the network $\epsilon_{\theta}$ with given $z_t$ and t: $\hat{\epsilon}=\epsilon_{\theta}(z_t;t)$.
We re-parameterize the above generative denoising process to adapt low-quality latent space enhancement in segmentation, as:
\begin{equation}
    \hat{z}_H = \mathrm{GLE}(z_L)= \frac{z_L-\sqrt{1-\overline{\alpha}_T}\epsilon_{\theta}(z_L; T)}{\sqrt{\overline{\alpha}_T}},
    \label{eq: denoise}
\end{equation}
where we consider low-quality feature $x_L$ as the noised feature and perform one-step denoising at the \textit{T}-th diffusion timestep.
The denoised output $\hat{z}_H$ is expected to be closer to the features extracted from clear images $z_H$. 
This single-step process significantly reduces computational overhead, making it more efficient when applied to segmentation models. 
After that, with $\hat{z}_H$ as input, the mask decoder can predict more precise masks, as: $m_p=\mathcal{D}_{\theta}(\hat{z}_H)$.

\subsection{Feature Distribution Alignment}
\label{sec: faw}
We employ the pre-trained U-Net in LDM as the denoising backbone. 
However, a significant challenge arises due to the substantial difference between the latent spaces in the original LDM (encoded by VAE) and segmentation models, leading to several technical issues for our application.

Firstly, there is a distribution gap between the two spaces and directly feeding segmentation features into U-Net may prevent it from fully exerting its denoising capabilities, as shown in the right part of Fig. \ref{fig:pipeline}.
To address this gap, we introduce a Feature Distribution Alignment (FDA) technique. Specifically, we add an adaptation weight $\gamma$ to scale the segmentation features, adjusting their variance to align more closely with the VAE's latent space.
This adjustment ensures that the features are compatible with U-Net’s optimal input space, improving the robustness and accuracy of the semantic interpretation and enhancing the denoising capability.
The LQ feature denoising process in Eq. \ref{eq: denoise} can be updated as:
\begin{equation}
    \hat{z}_H = \mathrm{GLE}(z_L)= \frac{\gamma z_L-\sqrt{1-\overline{\alpha}_T}\epsilon_{\theta}(\gamma z_L; T)}{\gamma \sqrt{\overline{\alpha}_T}},
    \label{eq: denoise-2}
\end{equation}
where we divide by $\gamma$ to restore its original distribution.
We experimentally verified in Sec. \ref{sec: ablation} that this simple operation effectively improves U-Net's denoising performance when applied to segmentation features.

\subsection{Channel Expansion for Head-tail Layers}
\label{sec: REI}
Another technical issue arises from the channel mismatch of the head and tail layers between the pre-trained U-Net and the segmentation features.  The U-Net in LDMs is designed for 4-channel input and output ($h \times w \times 4$), which does not match the dimension of SAM's latent space ($h\times w \times 256$).
We explore various methods to solve this problem (in Sec. \ref{sec: ablation}) and empirically find that fine-tuning new head and tail layers or an encoder-decoder for segmentation features is ineffective. This is likely due to difficulties in aligning with the pre-trained model's parameters while preserving its generalization ability.
To address this, we propose a Channel Replication and Expansion (CRE) method that replicates and concatenates the pre-trained weights of head and tail layers to match the required channel dimension. 
During training, the parameters of the head and tail layers remain frozen, while learnable LoRA layers are added to adapt to segmentation features. This approach effectively preserves the pre-trained generalization capacity while minimizing the number of learnable parameters.

\subsection{Training Method}
\label{sec: ft}
We employ a two-step fine-tuning process. In the first step, we fine-tune the denoising U-Net to reconstruct high-quality features. In the second step, we fine-tune the decoder with the restored features to further align the feature space for more accurate segmentation. 

\noindent\textbf{U-Net finetuning.} To adapt the pre-trained U-Net to the segmentation framework while preserving its inherent generalization ability, we employ the LoRA \cite{lora} scheme to fine-tune all the attention layers in the U-Net. During this step, we freeze the pre-trained image encoder and U-Net layers and only fine-tune the added LoRA layers. The estimated feature is compared with the corresponding HQ feature $z_H$ by a reconstruction loss, as:
\begin{equation}
    \mathcal{L}_{\mathrm{Rec}} = \mathcal{L}_{\mathrm{MSE}}(\mathrm{GLE}(z_L), z_H).
\end{equation}
This step significantly enhances performance without fine-tuning SAM’s parameters.

\noindent\textbf{Decoder finetuning.} Next, we use the reconstructed high-quality features to fine-tune the mask decoder for more precise segmentation.  Our experiments demonstrate that fine-tuning either the entire decoder or only the output tokens with these features further improves segmentation accuracy while maintaining generalization on clear images.
Focal Loss and Dice Loss are employed as segmentation loss functions, as:
\begin{equation}
    \mathcal{L}_{\mathrm{Seg}} = \mathcal{L}_{\mathrm{Dice}}(m_p, m_g) + \mathcal{L}_{\mathrm{Focal}}(m_p, m_g),
\end{equation}
where $m_p$ and $m_g$ indicate predicted and ground-truth masks respectively.

\section{Low-Quality Image Segmentation Dataset}
\label{sec:lq-dataset}
We construct a comprehensive low-quality image segmentation dataset dubbed \textbf{\textit{LQSeg}} that encompasses more complex and multi-level degradations, rather than relying on a single type of degradation for each image. The dataset is composed of images from several existing datasets with our synthesized degradations.
In this section, we first model a multi-level degradation process of low-quality images and then detail the dataset composition.

\subsection{{Multi-level Degradation Modeling}}
To model a more practical and complex degradation process, inspired by the previous work in image reconstruction \cite{realesrgan, bsrgan}, we utilize a mixed degradation method. 
Specifically, the degraded process is modeled as the random combination of the four common degradation models, including \textit{Blur}, \textit{Random Resize}, \textit{Noise}, and \textit{JPEG Compression}. 
Each degradation model encompasses various types, such as Gaussian and Poisson noise for \textit{Noise}, ensuring the diversity of the degradation process. 

To enrich the granularity of degradation,  we employ multi-level degradation by adjusting the downsampling rates. We employed three different resize rates, \textit{i.e.}, [1, 2, 4], which correspond to three degradation levels from slight to severe: LQ-1, LQ-2, and LQ-3. More implementation details are illustrated in \textit{Supplementary Material}.

\subsection{Dataset Composition}
Based on the above multi-level degradation model, we construct \textbf{\textit{LQSeg}} to train our model and evaluate the segmentation performance on different levels of low-quality images.
The images in LQSeg are sourced from several well-known existing datasets in the community with our synthesized degradation. 
In detail, for the \textbf{\textit{training set}}, we utilize the entire training sets of LVIS \cite{lvis}, ThinObject-5K \cite{thin}, and MSRA10K \cite{msra} as the source data and procedurally  
synthesize corresponding low-quality images. 
The \textbf{\textit{evaluation set}} is sourced from four subsets, \textit{i.e.}, 1) the whole test sets of ThinObject-5K and LVIS (seen sets), and 2) ECSSD \cite{ecssd} and COCO-val \cite{coco} (unseen sets). For each source image, We systematically generate three levels of degraded images to thoroughly assess the model's robustness. 

\begin{table*}[ht] \small
\centering
\renewcommand\arraystretch{1}
\setlength{\abovecaptionskip}{0pt}   
\setlength{\belowcaptionskip}{5pt}
\setlength{\tabcolsep}{1.8 mm}{
\begin{tabular}{ccccccc|cccccc}\toprule[1.pt]
     \multirow{3}{*}{{Method}} & \multicolumn{6}{c|}{ThinObject-5K} & \multicolumn{6}{c}{LVIS} \\
     & \multicolumn{2}{c}{{LQ-3}} & \multicolumn{2}{c}{{LQ-2}} &\multicolumn{2}{c|}{{LQ-1}} & \multicolumn{2}{c}{{LQ-3}} & \multicolumn{2}{c}{{LQ-2}} &\multicolumn{2}{c}{{LQ-1}} \\
\cmidrule(r){2-3}  \cmidrule(r){4-5} \cmidrule(r){6-7} \cmidrule(r){8-9}  \cmidrule(r){10-11} \cmidrule(r){12-13}
                               & IoU & Dice & IoU & Dice 
                            & IoU & Dice & IoU & Dice & IoU & Dice 
                            & IoU & Dice \\
\toprule[1.pt]
SAM   & 0.6285 & 0.7286  & 0.7054 &0.7939 & 0.7527 & 0.8343 & 0.4041  & 0.5005 &0.4886 & 0.5838 & 0.5325 & 0.6282 \\
RobustSAM  & 0.7015 & 0.7965 & 0.7648  & 0.8463 &0.7922 & \underline{0.8658} & 0.4517  & 0.5670 & 0.4962 & 0.6079 & 0.5262   & 0.6356 \\
PromptIR-SAM & 0.6380 & 0.7374 &0.7146 & 0.8018 & 0.7434 & 0.8248 & 0.4020 & 0.4978  & 0.4705 & 0.5677 & 0.5222 & 0.6187 \\
DiffBIR-SAM  & \underline{0.7055} & \underline{0.7917} & \underline{0.7772} & \underline{0.8531} & \underline{0.7927} & 0.8652  & \underline{0.5316} & \underline{0.6307} & \underline{0.5812} & \underline{0.6913} & \underline{0.6021} & \underline{0.7090} \\
\rowcolor{lightgray!40}
\textbf{GleSAM (Ours)} & \textbf{0.7594} & \textbf{0.8413}  & \textbf{0.8033}  & \textbf{0.8738} & \textbf{0.8277} & \textbf{0.8920} & \textbf{0.5535} & \textbf{0.6756} & \textbf{0.5916}  & \textbf{0.7066} & \textbf{0.6131}  & \textbf{0.7244} \\
\hline
\hline
SAM2  &0.7174 &0.8000 &0.7636 & 0.8373 & 0.7839 &  0.8536  & 0.5118 &0.6174 & 0.5634 &0.6633 & 0.6024 & 0.7004  \\
PromptIR-SAM2 & 0.7119 & 0.7945 & 0.7753 &0.8477 & 0.7801 & 0.8487 & 0.5017 &  0.6084 & 0.5529 & 0.6546 & 0.5875 & 0.6865 \\
DiffBIR-SAM2 & \underline{0.7348}  & \underline{0.8117} & \underline{0.7832} & \underline{0.8505} & \underline{0.7974} & \underline{0.8604} & \underline{0.5651} & \underline{0.6664} & \underline{0.6004} & \underline{0.7032} & \underline{0.6278} &  \underline{0.7380} \\
\rowcolor{lightgray!40}
\textbf{GleSAM2 (Ours)} & \textbf{0.7947} & \textbf{0.8653} & \textbf{0.8300} & \textbf{0.8896} &\textbf{0.8527} & \textbf{0.9072} & \textbf{0.5738} & \textbf{0.6887} & \textbf{0.6082} & \textbf{0.7161} & \textbf{0.6361} & \textbf{0.7402} \\
\toprule[1.pt]
\end{tabular}}
\caption{Performance comparison on the test set of Thinobject-5K \cite{thin} and LVIS \cite{lvis} datasets (seen datasets) with different levels of degradation. From LQ-1 to LQ-3, the degree of degradation increases progressively. We report IoU and Dice for comparison. Our GleSAM and GleSAM2 consistently outperform other competitors, especially on the most challenging LQ-3 version.
The words with boldface indicate the best results and those underlined indicate the second-best results.}
\vspace{-6pt}
\label{tab:thin}
\end{table*}

\begin{table*}[ht] \small
\centering
\renewcommand\arraystretch{1}
\setlength{\abovecaptionskip}{0pt}    
\setlength{\belowcaptionskip}{5pt}
\setlength{\tabcolsep}{1.8 mm}{
\begin{tabular}{ccccccc|cccccc}\toprule[1.pt]
     \multirow{3}{*}{{Method}} & \multicolumn{6}{c|}{{ECSSD}} & \multicolumn{6}{c}{{COCO}} \\ 
     & \multicolumn{2}{c}{{LQ-3}} & \multicolumn{2}{c}{{LQ-2}} & \multicolumn{2}{c|}{{LQ-1}} & \multicolumn{2}{c}{{LQ-3}} & \multicolumn{2}{c}{{LQ-2}}  & \multicolumn{2}{c}{{LQ-1}} \\ 
     \cmidrule(r){2-3}  \cmidrule(r){4-5} \cmidrule(r){6-7}  \cmidrule(r){8-9}  \cmidrule(r){10-11} \cmidrule(r){12-13}
                               & IoU  & Dice & IoU & Dice & IoU  & Dice & IoU & Dice & IoU  & Dice & IoU & Dice \\
\toprule[1.pt]
SAM   &0.5219 &0.6383  & 0.6156 & 0.7202 & 0.6778 & 0.7735 & 0.4048 & 0.5014 & 0.4848 & 0.5810 & 0.5392 & 0.6350 \\
RobustSAM  & 0.6535 &  \underline{0.7659} & 0.7271 &0.8240  & \underline{0.7758} & \underline{0.8606} & 0.4541  & 0.5711  & 0.4956 &0.6087 &0.5304 & 0.6401 \\
PromptIR-SAM & 0.5277 & 0.6442 & 0.6109 &0.7188 & 0.6752 &0.7726  & 0.4006  & 0.4969 & 0.4784 & 0.5753 & 0.5257 & 0.6219 \\
DiffBIR-SAM & \underline{0.6675} & 0.7652 & \underline{0.7408} & \underline{0.8244} & 0.7737 & 0.8501 & \underline{0.5260} & \underline{0.6263} & \underline{0.5742} & \underline{0.6918} & \underline{0.6112} & \underline{0.7080}\\
\rowcolor{lightgray!40} 
\textbf{GleSAM (Ours)} & \textbf{0.7332}  & \textbf{0.8282} &  \textbf{0.7944} & \textbf{0.8719} & \textbf{0.8257} & \textbf{0.8931} & \textbf{0.5495} & \textbf{0.6705} & \textbf{0.5900}& \textbf{0.7055} & \textbf{0.6158}& \textbf{0.7257}\\
\hline
\hline
SAM2 &0.6534 & 0.7540 & 0.7292 &0.8157 & 0.7625 & 0.8417 & 0.5176 &0.6230 & 0.5681 & 0.6692 &0.6018 &0.7000 \\
PromptIR-SAM2 & 0.6490  & 0.7526  & 0.7175  &0.8082 & 0.7519 & 0.8351 & 0.5088 &0.6153  & 0.5579 &0.6600 & 0.5905 & 0.6914\\
DiffBIR-SAM2 & \underline{0.7158} & \underline{0.8026} & \underline{0.7853} & \underline{0.8588} & \underline{0.8070} & \underline{0.8745}  & \underline{0.5680}  & \underline{0.6711} & \underline{0.6109} & \underline{0.7190} & \underline{0.6245} & \underline{0.7348} \\
\rowcolor{lightgray!40} 
\textbf{GleSAM2 (Ours)} & \textbf{0.7335} & \textbf{0.8258} & \textbf{0.7976} & \textbf{0.8726} & \textbf{0.8175} & \textbf{0.8853} & \textbf{0.5724} & \textbf{0.6867} & \textbf{0.6165} & \textbf{0.7246} & 
\textbf{0.6337}& \textbf{0.7371} \\
\toprule[1.pt]
\end{tabular}}
\caption{Zero-shot performance comparison on the ECSSD \cite{ecssd} and COCO \cite{coco} datasets (unseen datasets) with different levels of degradation. These results indicate that GleSAMs possesses significant robustness in zero-shot segmentation across different levels of degradations.}
\vspace{-15pt}
\label{tab:unseen-sets}
\end{table*}

\begin{figure*}[t]
  \centering
   \includegraphics[width=0.97\linewidth]{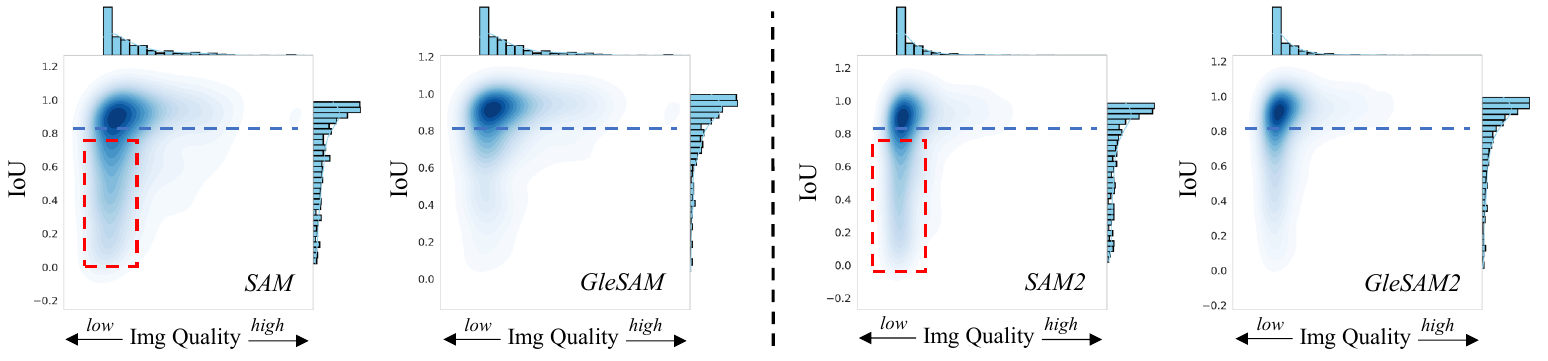}
   \caption{{Density distribution maps about IoU and image quality across different methods, including SAM, GleSAM, SAM2, and GleSAM2. The image quality is calculated using the Laplacian operator in OpenCV. The red dashed box highlights the area where our method demonstrates improved segmentation performance compared to SAM, particularly in lower-quality images.} }
   \vspace{-10pt}
   \label{fig:dense-map}
\end{figure*}

\section{Experiment}
We conduct extensive experiments to verify our method across images of varying quality. All proposed techniques can be applied to SAM and SAM2, referred to as GleSAM and GleSAM2. In practice, our models perform well on low-quality (Tab. \ref{tab:thin}, \ref{tab:unseen-sets}) and clear images (Tab. \ref{tab:ablation-ft}) and they generalize effectively to unseen degradations (Tab. \ref{tab:robsutseg}).

\subsection{Experimental Setup}
\textbf{Implement Details.}
Our model is trained using the AdamW optimizer with the learning rate of $1\times10^{-4}$  and batch size of 4. The pre-trained U-Net in Stable Diffusion (SD) 2.1-base \cite{ldm} is adopted as the denoising backbone.
We use 3 clicks as SAM's prompts by default.
Our approach can be efficiently trained on 4$\times$ A100 GPUs within approximately 30 hours, during which we fine-tune the U-Net for 100K iterations and the decoder for only 20K iterations.

\noindent\textbf{Evaluation Metrics.} We employ three metrics to assess our model's performance, including  Intersection over Union (IoU), Dice Coefficient (Dice), and Pixel Accuracy (PA).

\subsection{Performance Comparisons}
In this experiment, we evaluate the performance on the test set of our LQSeg, including seen-set (Tab. \ref{tab:thin}) and unseen-set (Tab. \ref{tab:unseen-sets}) evaluations.  
We compare our method with a set of comparison baselines to quantify the performance gains.
For SAM-based comparisons, besides the original SAM \cite{sam}, we also compare with the RobustSAM \cite{robustsam}, which has improved robustness on the degraded dataset. 
Additionally, we compare with two-stage methods, \textit{i.e.}, reconstructing images first with image reconstruction (IR) networks and passing the restored clear images to the SAM and SAM2. We use two state-of-the-art IR networks for comparison: PromptIR \cite{promptir}, and diffusion-based DiffBIR \cite{diffbir}.

\noindent\textbf{Performance Comparison on Seen Datasets.}
In Tab. \ref{tab:thin}, we evaluate the performance of our GleSAM and GleSAM2 on two seen datasets: ThinObject-5K \cite{thin} and LVIS \cite{lvis}. 
Each dataset contains three levels of degradation. Our method demonstrates superior performance across all degradation levels, effectively handling low-quality images.  As the degradation level increases (from LQ-1 to LQ-3), GleSAMs show increasingly significant performance improvements compared to the baseline, highlighting its robustness against challenging degradations.

\noindent\textbf{Performance Comparison on Un-seen Datasets.}
In Tab. \ref{tab:unseen-sets}, we evaluate the zero-shot segmentation performance of GleSAMs on two unseen datasets: ECSSD \cite{ecssd} and COCO \cite{coco}, all of which are synthesized with different levels of degradations. GleSAM consistently outperforms other methods, particularly on the most challenging LQ-3 version, underscoring its strong zero-shot generalization capabilities and potential for real-world applications. To further assess segmentation quality, we plot the density distribution maps of image quality and segmentation IoU in Fig. \ref{fig:dense-map}. Compared to the baselines, our method achieves overall improvement and more stable performance on low-quality images, especially for those of inferior quality.

\noindent\textbf{Validation with Other Degradations.}
To validate the model's generalization on other unseen degradations, we evaluated GleSAM and GleSAM2 on RobustSeg-style degradations \cite{robustsam}, with the results presented in Tab. \ref{tab:robsutseg}. 
These degradations were not used during training. Our method consistently outperforms SAM and SAM2 and even surpasses RobustSAM which is specifically trained on RobustSeg. This demonstrates the strong generalization of our method across diverse degradations.

\begin{table}[t] \small
\centering
\renewcommand\arraystretch{1}
\setlength{\abovecaptionskip}{0pt}    
\setlength{\belowcaptionskip}{5pt}
\setlength{\tabcolsep}{2.5 mm}{
\begin{tabular}{ccccc}\toprule[1.pt]
     \multirow{2}{*}{{Method}} &\multicolumn{2}{c}{{ECSSD}} &\multicolumn{2}{c}{{COCO}} \\
\cmidrule(r){2-3}  \cmidrule(r){4-5} 
                            & IoU & Dice  & IoU & Dice \\
\toprule[1.pt]
SAM \cite{sam} & 0.7833 & 0.8573 & 0.6512 & 0.7457 \\
RobustSAM \cite{robustsam}  & 0.8427 & 0.9044 &0.6543 &0.7518 \\
\rowcolor{lightgray!40} 
\textbf{GleSAM (Ours)}  &\textbf{0.8568} & \textbf{0.9119} & \textbf{0.6678} & \textbf{0.7700} \\
\hline
SAM2  & 0.8211 & 0.8817 & 0.6817 & 0.7753\\
\rowcolor{lightgray!40} 
\textbf{GleSAM2 (Ours)} &\textbf{0.8611}  & \textbf{0.9133} &\textbf{ 0.6869 }& \textbf{0.7836} \\
\toprule[1.pt]
\end{tabular}}
\caption{Zero-shot performance comparison on RobustSeg-style \cite{robustsam}  degradations. Performance is tested on the unseen ECSSD and COCO datasets. Note that our methods are not trained on such degradations. The superior performance of our method demonstrates robustness against various forms of degradation.}
\vspace{-5pt}
\label{tab:robsutseg}
\end{table}

\begin{table}[t] \small
\centering
\renewcommand\arraystretch{1}
\setlength{\abovecaptionskip}{0pt}
\setlength{\belowcaptionskip}{5pt}
\setlength{\tabcolsep}{2 mm}{
\begin{tabular}{lcccc}\toprule[1.pt]
     \multirow{2}{*}{{Method}} &\multicolumn{2}{c}{{ECSSD}} &\multicolumn{2}{c}{{COCO}}\\
     \cmidrule(r){2-3}  \cmidrule(r){4-5} 
     & IoU & Dice & IoU & Dice \\
\toprule[1.pt]
Baseline & 0.6054 & 0.7107 & 0.4763 & 0.5725 \\
+ Gle \& CRE  & 0.6567 & 0.7657 & 0.4958 & 0.5963  \\
\rowcolor{lightgray!40} 
+ Gle \& CRE \& FDA  & \textbf{0.7104} & \textbf{0.8020} & \textbf{0.5166} & \textbf{0.6174} \\
\toprule[1.pt]
\end{tabular}}
\caption{Ablation study of each component in the proposed method, evaluated on the unseen ECSSD and COCO datasets. Each additional component positively affects the performance, demonstrating the effectiveness of the proposed methods. }
\vspace{-15pt}
\label{tab:ablation}
\end{table}

\begin{table}[t] \small
\centering
\renewcommand\arraystretch{1.}
\setlength{\abovecaptionskip}{0pt}    
\setlength{\belowcaptionskip}{5pt}
\setlength{\tabcolsep}{1.1 mm}{
\begin{tabular}{ccccccc}\toprule[1.pt]
     \multirow{2}{*}{{Method}} & \multicolumn{2}{c}{{LQ}} &\multicolumn{2}{c}{{Clear}} & \multicolumn{2}{c}{{Average}}\\
\cmidrule(r){2-3}  \cmidrule(r){4-5} \cmidrule(r){6-7}
                               & IoU & Dice & IoU & Dice & IoU & Dice 
                             \\
\toprule[1.pt]
\multicolumn{5}{l}{\textbf{\textit{w/o Fine-tuning SAM:}}} \\
SAM & 0.5407 & 0.6416 & 0.7830  & 0.8554 & 0.6619 & 0.7485\\
\rowcolor{lightgray!40} 
Ours & \textbf{0.6135} &  \textbf{0.7097} & \textbf{0.7846}   &\textbf{0.8567} & \textbf{0.6991} & \textbf{0.7832} \\
\midrule[0.8pt]
\multicolumn{5}{l}{\textbf{\textit{Fine-tuning SAM: }}} \\
SAM-FT-T & 0.6305 & 0.7374 & 0.5847 & 0.7029 & 0.6076 & 0.7202 
\\
SAM-FT-D  & 0.6327 & 0.7385  & 0.6071 & 0.7242 & 0.6199 & 0.7314\\
\rowcolor{lightgray!40} 
Ours-FT-T & \underline{0.6759} & \underline{0.7751} & \textbf{0.8061} & \textbf{0.8747} & \underline{0.7410} & \underline{0.8249}\\
\rowcolor{lightgray!40} 
Ours-FT-D & \textbf{0.6848} & \textbf{0.7825} &  \underline{0.8022} & \underline{0.8704} & \textbf{0.7435} &\textbf{ 0.8265} \\
\toprule[1.pt]
\end{tabular}}\caption{Effect of Fine-tuning SAM. The performance is evaluated on the unseen ECSSD and COCO datasets. ``FT-T'' and ``FT-D'' indicate fine-tuning the SAM's mask token and decoder respectively. ``LQ'' indicates the mean performance on three levels of degraded data, and ``Clear'' indicates the results on the original clear images.}
\vspace{-5pt}
\label{tab:ablation-ft}
\end{table}

\subsection{Ablation Study}
We conduct ablation experiments in Tab. \ref{tab:ablation}, \ref{tab:ablation-ft} to further understand the impact of each component of our framework. 

\begin{figure}[t]
  \centering
   \includegraphics[width=1\linewidth]{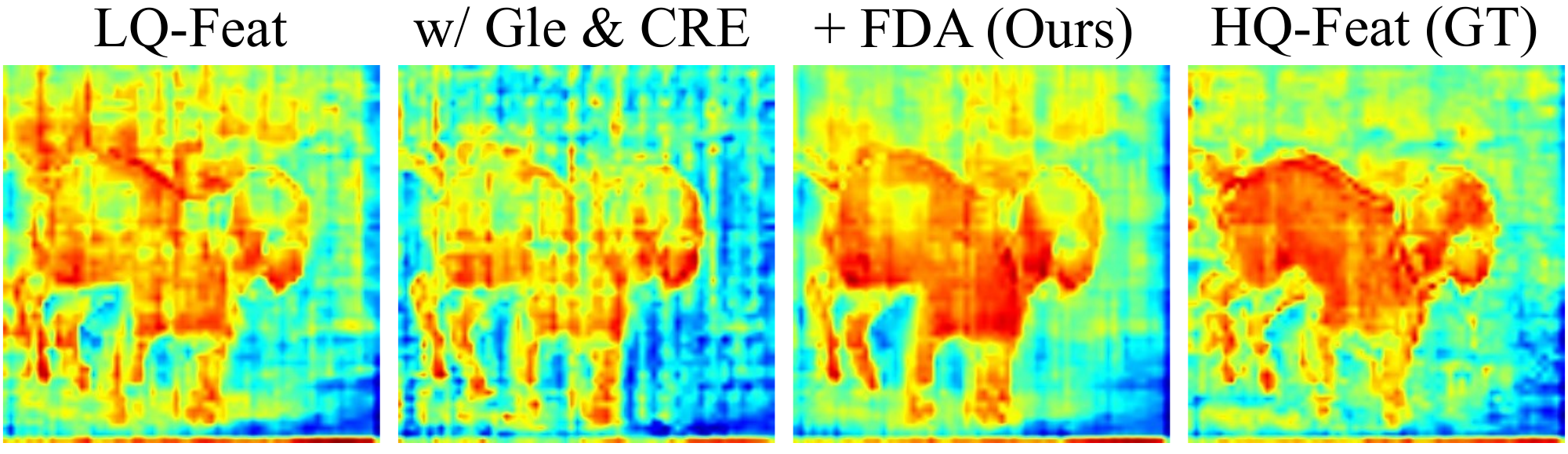}
   \caption{Qualitative visualization of the enhanced latent features. The clearest feature is obtained when combining all modules.}
   \label{fig:ablation-vis}
   \vspace{-10pt}
\end{figure}

\noindent\textbf{Effect of Each Component.}
In Tab. \ref{tab:ablation}, we validated the effectiveness of each proposed module in GleSAM. ``Gle'' indicates our framework with generative latent space enhancement. The results show that each introduced method significantly improves the performance. We also make a qualitative visualization in Fig. \ref{fig:ablation-vis}, which shows that the clearest latent feature is obtained when combining all modules. Additionally, in Tab. \ref{tab:ablation-ft}, without fine-tuning the decoder, our method improves the results on LQ images by about 7 points, while also preserving the robustness of SAM on clear conditions, indicating the robustness of our method for both degraded and clear images.

\noindent\textbf{Effect of Fine-tuning SAM.}  We explore two common configurations to fine-tune SAM: fine-tuning the entire SAM's decoder and the output mask token. Our experiments reveal that directly fine-tuning SAM's decoder or output mask token on degraded images leads to a significant drop in zero-shot performance on clear images, with the IoU decreasing by nearly \textbf{\textit{20 points}}.  
In contrast, our method further improves performance on both low-quality and clear images by fine-tuning the entire decoder or mask token, enabling the segmentation of images with any quality.

\subsection{Detailed Analysis}
\label{sec: ablation}

\begin{figure}[t]
  \centering
   \includegraphics[width=1\linewidth]{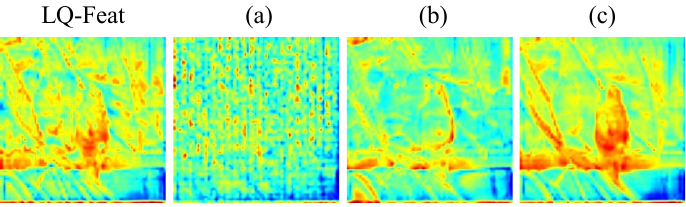}
   \caption{Qualitative visualization of the enhanced latent features generated by different channel expansion methods. Our proposed CRE method (c) produces more salient features.}
   \label{fig:ablation-vis-cre}
   \vspace{-10pt}
\end{figure}

\noindent\textbf{Analysis of channel expansion methods.}
To address the issue of channel mismatch between SAM and the pre-trained U-Net, we explored various strategies including (a) using two simple convolutional layers to reduce and expand the channels of segmentation features as needed, (b) fine-tuning new head and tail layers from scratch, and (c) our CRE method. 
Our results (in Tab. \ref{tab:ablation-REI} and Fig. \ref{fig:ablation-vis-cre}) show that strategies (a) and (b) are ineffective, likely because the new layers couldn’t leverage the pre-trained knowledge. In contrast, our method effectively resolves this issue by replicating pre-trained parameters and fine-tuning with LoRA.

\begin{figure}[t]
  \centering
   \includegraphics[width=1.\linewidth]{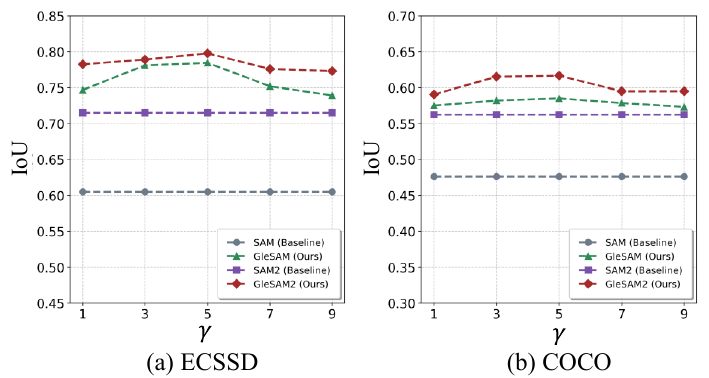}
   \caption{Ablation study of adaption weight $\gamma$. }
   \label{fig:ablation-k}
\end{figure}

\begin{table}[t] \small
\centering
\renewcommand\arraystretch{1}
\setlength{\abovecaptionskip}{0pt}    
\setlength{\belowcaptionskip}{5pt}
\setlength{\tabcolsep}{1.2 mm}{
\begin{tabular}{c|ccc}\toprule[1.pt]
     {Method} & IoU & Dice  & PA \\
\toprule[1.pt]
(a) Additional encoder and decoder & 0.4544 &  0.5842 &  0.6106 \\
(b) New head and tail layers & 0.6014 &0.7077 & 0.7782 \\
\rowcolor{lightgray!40}
(c) Replicate and Expansion (Ours) & \textbf{0.6567} & \textbf{0.7657} & \textbf{0.8400} \\
\toprule[1.pt]
\end{tabular}}
\caption{Analysis of the proposed CRE. It significantly outperforms alternative approaches, achieving higher scores.}
\vspace{-8pt}
\label{tab:ablation-REI}
\end{table}

\noindent\textbf{Analysis of hyperparameter $\gamma$.}
We use an adaption weight $\gamma$ to align the distribution of the latent space between LDM and SAM. 
To determine the optimal value of $\gamma$, we empirically test five different values on the ECSSD and COCO datasets. The results, shown in Fig. \ref{fig:ablation-k}, suggest that $\gamma = 5$ is the most effective setting, providing strong generalization across all models and datasets. Therefore, we adopt $\gamma = 5$ as the default value in our experiments.

\noindent\textbf{Analysis of LoRA ranks.}
We incorporate learnable LoRA layers to fine-tune the pre-trained denoising U-Net. 
Here, we evaluate the impact of different LoRA ranks on segmentation performance. The results are shown in Tab. \ref{tab:rank}. 
We tested the results and the corresponding number of learnable parameters for setting a rank to 4, 8, and 16. 
We found that setting the rank to 8 can obtain good results while maintaining an acceptable number of parameters.

\begin{table}[t] \small
\centering
\renewcommand\arraystretch{1.}
\setlength{\abovecaptionskip}{0pt}    
\setlength{\belowcaptionskip}{5pt}
\setlength{\tabcolsep}{2.2 mm}{
\begin{tabular}{cccccc}\toprule[1.pt]
     Training & \multicolumn{3}{c}{Ours} & \multirow{2}{*}{{LQ-RS}} & \multirow{2}{*}{{AVG}} \\
     \cmidrule(r){2-4}  
     Data & LQ-1 & LQ-2 & LQ-3  \\
\toprule[1.pt]
LQ-1 & 0.6981 & 0.6663  & 0.6076 & 0.7587 & 0.6827\\
LQ-2 & 0.7168 & 0.6843  & 0.6318  & \underline{0.7621} & 0.6988 \\
LQ-3 & \underline{0.7185}  & \underline{0.6910}  & \underline{0.6403} & 0.7619 & \underline{0.7030}  \\
\rowcolor{lightgray!40}
LQ-1,2,3 & \textbf{0.7208} & \textbf{0.6922} & \textbf{0.6414} & \textbf{0.7623} & \textbf{0.7042}  \\
\toprule[1.pt]
\end{tabular}}
\caption{Ablation study for degradation levels during training.  LQ-RS indicates the RobustSeg-style \cite{robustsam} degradation. AVG indicates the average performance. We report the IoU for comparison. The results on the unseen ECSSD and COCO datasets reveal that multi-level degradation contributes to more robust performance.}
\vspace{-5pt}
\label{tab:ablation-dataset}
\end{table}

\begin{table}[t] \small
\centering
\renewcommand\arraystretch{1.}
\setlength{\abovecaptionskip}{0pt}   
\setlength{\belowcaptionskip}{5pt}
\setlength{\tabcolsep}{3.5 mm}{
\begin{tabular}{ccccc}\toprule[1.pt]
     \multirow{2}{*}{{Rank}} & \multicolumn{3}{c}{{LQ}} & Learnable\\
\cmidrule(r){2-4}  
                               & IoU & Dice & PA & Params\\
\toprule[1.pt]
4  & 0.7760 & 0.8597 & 0.9434 & 16.25M \\
8  & \textbf{0.7844} & \textbf{0.8644} & \textbf{0.9456}  & 32.49M \\
16  & 0.7697 & 0.8543 & 0.9418 & 64.99M \\
\toprule[1.pt]
\end{tabular}}
\caption{Analysis of LoRA ranks in U-Net.}
\vspace{-5pt}
\label{tab:rank}
\end{table}

\begin{table}[t] \small
\centering
\renewcommand\arraystretch{1.}
\setlength{\abovecaptionskip}{0pt}
\setlength{\belowcaptionskip}{5pt}
\setlength{\tabcolsep}{2 mm}{
\begin{tabular}{cccccc}\toprule[1.pt]
     \multirow{2}{*}{{Method}}  & Learnable & Num. &  Training & Inference\\
    & Params    & GPUs &  Time (h) & Speed (s) \\
\toprule[1.pt]
SAM & 1250 MB & 256 & N/A  & 0.32 \\
\rowcolor{lightgray!40}
GleSAM & 47 MB & 4 & 30 & 0.38 (+0.06) \\
\toprule[1.pt]
\end{tabular}}
\caption{Computational requirements of GleSAM \textit{vs} SAM.}
\vspace{-13pt}
\label{tab:speed}
\end{table}

\noindent\textbf{Analysis of degradation levels.}
To investigate the necessity of training on all three degradation levels together, we train GleSAM on LQ datasets for each degradation level (LQ-1, LQ-2, LQ-3) individually.
We evaluate the models' performance on ECSSD and COCO datasets using three degraded levels and RobustSeg-style degradations. 
The results in Tab. \ref{tab:ablation-dataset} show that training on all three degradation levels together contributes to more robust performance across varying levels of image degradation.

\noindent\textbf{Analysis of computational requirements.}
In Tab. \ref{tab:speed}, we report detailed training and inference comparisons between our GleSAM and SAM. 
Although GleSAM demonstrates significantly improved robustness, it introduces only marginal learnable parameters and incurs a slight trade-off in inference speed. The additional parameters can be efficiently optimized in 30 hours on four A100 GPUs.

\section{Conclusion}
We present GleSAM, a solution to enhance Segment Anything Models (SAM and SAM2) for robust segmentation across images of any quality, particularly those with severe degradation. 
We incorporate the generative ability of pre-trained diffusion models into the latent space of SAMs to enhance low-quality features, thus promoting more robust segmentation.
Our approach is further supported by the LQSeg dataset, which includes diverse degradation types and levels, allowing for more comprehensive model training and evaluation. Extensive experiments demonstrate that GleSAM achieves superior performance on degraded images while maintaining generalization to clear ones, and it performs well on unseen degradations, highlighting its robustness and versatility. 
This work extends SAM capabilities, offering a practical approach for degraded scenarios.




\section{More Quantitative and Qualitative Results}
\label{sec:sm-exper}
In this section, we present additional experiments to validate the model's robustness across other datasets, backbones, and prompts. Moreover, we provide more qualitative visual analyses.

\subsection{Zero-shot performance comparison on BDD-100K dataset}
To comprehensively evaluate the robustness and effectiveness of our methods under real-world degradation scenarios, we extend our experiments to the BDD-100K dataset \cite{bdd100k}. This dataset poses significant challenges due to its extensive diversity of real-world degradation factors, including adverse weather conditions (\eg, rain, fog) and inconsistent lighting environments. 
These characteristics make it a critical benchmark for testing segmentation models' performance in practical, uncontrolled settings.
In this evaluation, we compare our proposed methods with several state-of-the-art approaches, including SAM \cite{sam}, RobustSAM \cite{robustsam}, and SAM2 \cite{sam2}, to assess their relative performance. 
The detailed results are presented in Table \ref{tab:bdd}, highlighting the superiority of our methods in handling real-world degradation scenarios.


\subsection{Comparison across Different Backbones.}
In Tab. \ref{tab:sm-backbone}, we conduct a thorough comparison between SAM and GleSAM across various ViT \cite{vit} backbones, including ViT-Base (B), ViT-Large (L), and ViT-Huge (H).  In Tab. \ref{tab:sm-backbone-sam2}, we conduct a thorough comparison between SAM2 and GleSAM2 across various Hiera \cite{hiera} backbones, including Hiera-Tiny (T), Hiera-Small (S), Hiera-Base (B), Hiera-Large (L).
We comprehensively access the models on the seen ECSSD \cite{ecssd} dataset. The performance of three degraded levels is reported.
These results demonstrate that GleSAM/GleSAM2 consistently outperforms SAM/SAM2 with significant margins on various sizes of backbones.

\begin{table}[t] \small
\centering
\renewcommand\arraystretch{1}
\setlength{\abovecaptionskip}{0pt}%
\setlength{\belowcaptionskip}{5pt}%
\setlength{\tabcolsep}{4 mm}{
\begin{tabular}{cccc}\toprule[1.pt]
     \multirow{2}{*}{{Method}} &\multicolumn{3}{c}{{BDD-100K}} \\
\cmidrule(r){2-4} 
                            & IoU & Dice  & PA \\
\toprule[1.pt]
SAM & 0.8650 &  0.9216  & 0.9911  \\
RobustSAM &0.8708 & 0.9218 & 0.9930 \\
\rowcolor{lightgray!40} 
\textbf{GleSAM (Ours)} &\textbf{0.8775} & \textbf{0.9238} & 0\textbf{.9933} \\
\hline
SAM2 & 0.8891 & 0.9369  &0.9922 \\
\rowcolor{lightgray!40} 
\textbf{GleSAM2 (Ours)} &\textbf{0.9087} & \textbf{0.9452}  & \textbf{0.9947} \\
\toprule[1.pt]
\end{tabular}}
\caption{Zero-shot performance comparison on BDD-100K dataset. The superior results highlight the superiority of our methods in handling real-world degradation scenarios.}
\label{tab:bdd}
\end{table}

\begin{table*}[h] \small
\centering
\renewcommand\arraystretch{1.25}
\setlength{\abovecaptionskip}{0pt}%
\setlength{\belowcaptionskip}{5pt}%
\setlength{\tabcolsep}{2 mm}{
\begin{tabular}{ccccccccccccc}\toprule[1.pt]
     \multirow{2}{*}{{Method}} &\multicolumn{3}{c}{{LQ-3}} &\multicolumn{3}{c}{{LQ-2}} & \multicolumn{3}{c}{{LQ-1}} & \multicolumn{3}{c}{{Average}} \\
\cmidrule(r){2-4}  \cmidrule(r){5-7} \cmidrule(r){8-10} \cmidrule(r){11-13} 
                            & IoU & Dice & PA & IoU & Dice & PA & IoU & Dice & PA & IoU & Dice & PA \\
\toprule[1.pt]
SAM-B & 0.5425 & 0.6706  &0.8110  & 0.6018 &0.7203 & 0.8538 & 0.6321 & 0.7452 & 0.8705 & 0.5921 & 0.7120 & 0.8451  \\
\rowcolor{lightgray!40} 
\textbf{GleSAM-B}  & \textbf{0.6843} & \textbf{0.7935}  & \textbf{0.9145}  & \textbf{0.7329}  & \textbf{0.8275}  & \textbf{0.9303} & \textbf{0.7694}  &\textbf{0.8553}  & \textbf{0.9404}  & \textbf{0.7289} & \textbf{0.8254} &\textbf{0.9284}  \\
SAM-L & 0.5219 &0.6383 & 0.7915 & 0.6156 & 0.7202 &  0.8803 & 0.6778 & 0.7735 & 0.9023 & 0.6051 & 0.7107 & 0.8580\\
\rowcolor{lightgray!40} 
\textbf{GleSAM-L} & \textbf{0.7332} & \textbf{0.8282} &\textbf{0.9291}  & \textbf{0.7944} & \textbf{0.8719} & \textbf{0.9491} & \textbf{0.8257} & \textbf{0.8931}& \textbf{0.9666} & \textbf{0.7844} & \textbf{0.8644} & \textbf{0.9483}\\
SAM-H & 0.5912 & 0.7061 & 0.8129 &  0.6738  &0.7743  &0.8714& 0.7333  &0.8221  &0.9069 & 0.6661 & 0.7675 & 0.8637  \\
\rowcolor{lightgray!40} 
\textbf{GleSAM-H}   & \textbf{0.6913}  & \textbf{0.7976}  & \textbf{0.9152}  & \textbf{0.7558}  & \textbf{0.8444}  & \textbf{0.9351} & \textbf{0.8011} & \textbf{0.8764}  & \textbf{0.9483} & \textbf{0.7494} & \textbf{0.8395} & \textbf{0.932}9 \\
\toprule[1.pt]
\end{tabular}}
\caption{Performance comparison between SAM and GleSAM across different backbones. ``B'', ``L'', and ``H'' indicate ViT-Base, Large, and Huge, respectively. GleSAM consistently achieves superior performance. }
\label{tab:sm-backbone}
\end{table*}

\begin{table*}[h] \small
\centering
\renewcommand\arraystretch{1.25}
\setlength{\abovecaptionskip}{0pt}%
\setlength{\belowcaptionskip}{5pt}%
\setlength{\tabcolsep}{2 mm}{
\begin{tabular}{ccccccccccccc}\toprule[1.pt]
     \multirow{2}{*}{{Method}} &\multicolumn{3}{c}{{LQ-3}} &\multicolumn{3}{c}{{LQ-2}} & \multicolumn{3}{c}{{LQ-1}} & \multicolumn{3}{c}{{Average}} \\
\cmidrule(r){2-4}  \cmidrule(r){5-7} \cmidrule(r){8-10} \cmidrule(r){11-13} 
                            & IoU  & Dice &PA  & IoU  & Dice &PA & IoU  & Dice &PA & IoU  & Dice &PA\\
\toprule[1.pt]
SAM2-T  & 0.6411 & 0.7477  &0.8486  &0.7182 & 0.8096 &  0.9011 & 0.7550 & 0.8385  &0.9197 & 0.7047 & 0.7986 & 0.8898 \\
\rowcolor{lightgray!40} 
\textbf{GleSAM2-T} & \textbf{0.6949}  & \textbf{0.7859}  & \textbf{0.9032}  & \textbf{0.7555}  &\textbf{0.8329}  &\textbf{0.9225} &\textbf{0.7898}  &\textbf{0.8591}  &\textbf{0.9365} & \textbf{0.7467} & \textbf{0.8260} & \textbf{0.9207} \\
SAM2-S & 0.6468  &0.7537 & 0.8838  &0.7313   &0.8227  &0.9243  &0.7592  &0.8423   &0.9331 & 0.7124 & 0.8062 & 0.9137 \\
\rowcolor{lightgray!40} 
\textbf{GleSAM2-S}  & \textbf{0.7234} &\textbf{0.8196} & \textbf{0.9269}  & \textbf{0.7840} & \textbf{0.8621}  & \textbf{0.9443} & \textbf{0.8090}  &\textbf{0.8789}  & \textbf{0.9506}  & \textbf{0.7721} & \textbf{0.8535} & \textbf{0.9406}\\
SAM2-B & 0.6535 & 0.7504 & 0.8753 & 0.7292 &0.8157& 0.9059  & 0.7625 &0.8417 & 0.9204 & 0.7151 &0.8026 & 0.9005 \\
\rowcolor{lightgray!40} 
\textbf{GleSAM2-B} & \textbf{0.7335} & \textbf{0.8258} & \textbf{0.9242} & \textbf{0.7976} & \textbf{0.8726} & \textbf{0.9465} & \textbf{0.8175} & \textbf{0.8853} & \textbf{0.9508} & \textbf{0.7812} & \textbf{0.8612} & \textbf{0.9405} \\
SAM2-L & 0.6458  &0.7475  &0.8475 &0.7348  &0.8207  &0.9067  &0.7732  &0.8503  &0.9262 & 0.7179 & 0.8061 &0.8934\\
\rowcolor{lightgray!40} 
\textbf{GleSAM2-L} & \textbf{0.7244}  & \textbf{0.8214}  & \textbf{0.9233} & \textbf{0.7900} & \textbf{0.866}3  & \textbf{0.9434} & \textbf{0.8170}  & \textbf{0.8848}  & \textbf{0.9522} & \textbf{0.7771} & \textbf{0.8575} & \textbf{0.9525} \\
\toprule[1.pt]
\end{tabular}}
\caption{Performance comparison between SAM2 and GleSAM2 across different backbones. ``T'', ``S'', ``B'' and ``L'' indicate Hiera-Tiny, Small, Base, and Large, respectively. GleSAM2 consistently achieves superior performance. }
\label{tab:sm-backbone-sam2}
\end{table*}

\begin{table*}[!h] \small
\centering
\renewcommand\arraystretch{1.25}
\setlength{\abovecaptionskip}{0pt}%
\setlength{\belowcaptionskip}{5pt}%
\setlength{\tabcolsep}{1.8 mm}{
\begin{tabular}{ccccccc|cccccc}\toprule[1.pt]
     \multirow{3}{*}{{Method}} & \multicolumn{6}{c|}{{\textbf{\textit{GT-Box}}}} & \multicolumn{6}{c}{{\textbf{\textit{Noise-box}}}} \\ 
     & \multicolumn{2}{c}{{LQ-3}} & \multicolumn{2}{c}{{LQ-2}} & \multicolumn{2}{c|}{{LQ-1}} & \multicolumn{2}{c}{{LQ-3}} & \multicolumn{2}{c}{{LQ-2}}  & \multicolumn{2}{c}{{LQ-1}} \\ 
     \cmidrule(r){2-3}  \cmidrule(r){4-5} \cmidrule(r){6-7}  \cmidrule(r){8-9}  \cmidrule(r){10-11} \cmidrule(r){12-13}
                               & IoU  & Dice & IoU & Dice & IoU  & Dice & IoU & Dice & IoU  & Dice & IoU & Dice \\
\toprule[1.pt]
SAM   &0.7066 & 0.8109 & 0.7634 & 0.8522 & 0.7993 & 0.8771 & 0.6250 & 0.7400 & 0.6592 & 0.7659 & 0.6824 & 0.7821 \\
\rowcolor{lightgray!40} 
\textbf{GleSAM (Ours)} & \textbf{0.7718}  & \textbf{0.8575} &  \textbf{0.8316} & \textbf{0.9004} & \textbf{0.8605} & \textbf{0.9194} & \textbf{0.6845} & \textbf{0.7839} & \textbf{0.7284}& \textbf{0.8172} & \textbf{0.7547}& \textbf{0.8362} \\
\hline
\hline
SAM2 &0.7816 & 0.8667 & 0.8312 &0.8999 & 0.8501 & 0.9125 & 0.6818 &0.7855 & 0.7255 & 0.8194 &0.7421 &0.8313 \\
\rowcolor{lightgray!40}
\textbf{GleSAM2 (Ours)} & \textbf{0.8185} & \textbf{0.8928} & \textbf{0.8644} & \textbf{0.9225} & \textbf{0.8857} & \textbf{0.9358} & \textbf{0.7028}  & \textbf{0.8111 }& \textbf{0.7346}  & \textbf{0.8327}  & \textbf{0.7498}  & \textbf{0.8438}  \\ 
\toprule[1.pt]
\end{tabular}}
\caption{Performance comparison under different prompts. We use GT-Box and Noise-Box as prompts.  The GT-Box is generated based on the GT-mask, while the Noise-Box is obtained by adding noise to the GT-Box with the noise-scale of 0.2, following \cite{dndetr}.  We present results on three different quality degraded datasets from ECSSD, demonstrating the robustness of our method under various prompts.}
\label{tab:sm-box-prompt}
\end{table*}

\begin{figure*}[h]
  \centering
   \includegraphics[width=1\linewidth]{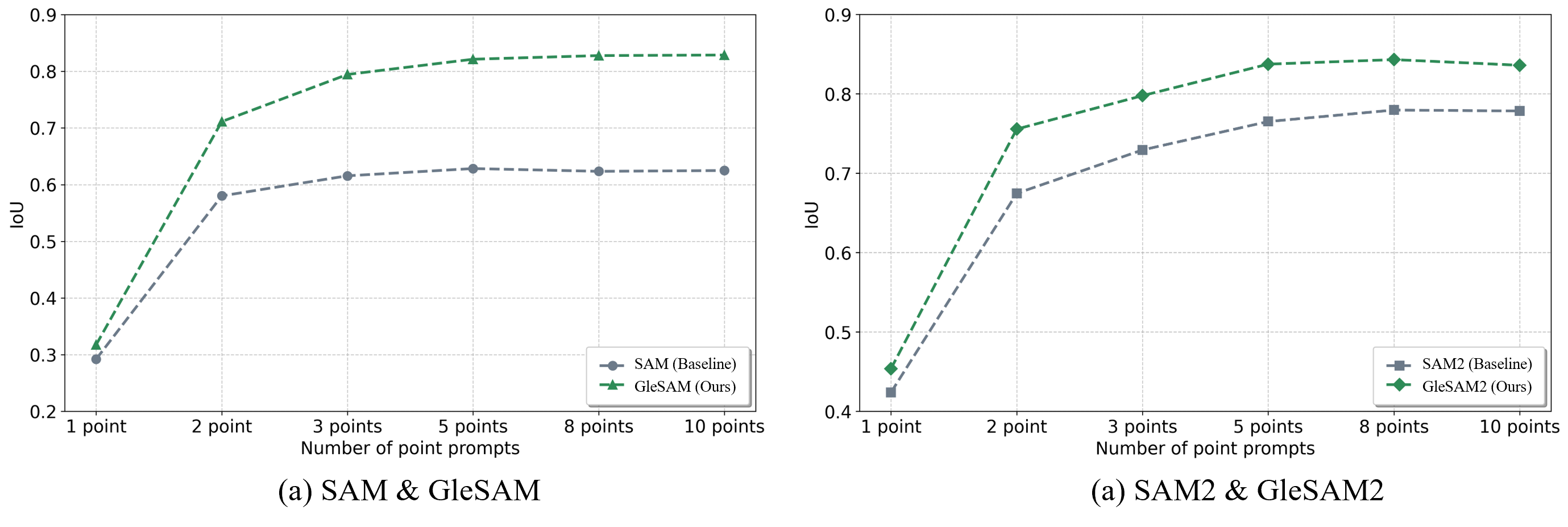}
   \caption{Performance comparison of interactive segmentation with varying quantities of input points on the unseen ECSSD dataset. GleSAMs consistently outperform SAMs across a range of point counts, demonstrating a more significant improvement. }
   \label{fig:sm-inter-seg}
\end{figure*}

\begin{figure*}[h]
  \centering
   \includegraphics[width=1\linewidth]{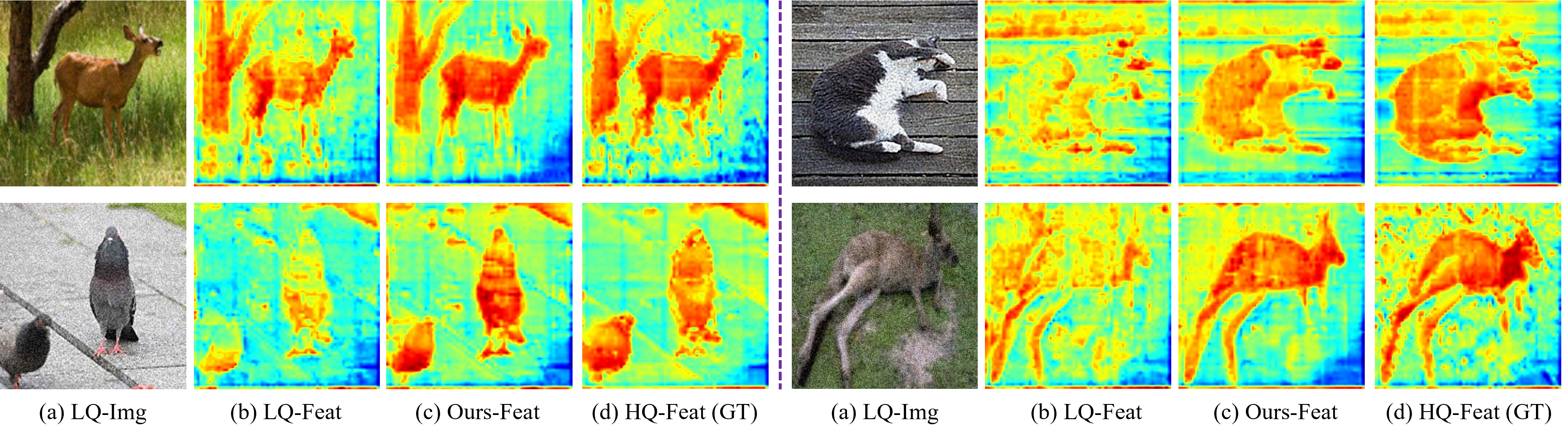}
   \caption{Visualization of feature maps for low-quality images (LQ-Feat), high-quality images (HQ-Feat), and features reconstructed by our method (Ours-Feat). Compared to LQ-Feat, our method effectively reconstructs features, aligning them closely with HQ-Feat, demonstrating the capability to restore high-quality representations from degraded inputs.}
   \label{fig:sm-lqfeat}
\end{figure*}

\begin{figure*}[t]
  \centering
\includegraphics[width=0.9\linewidth]{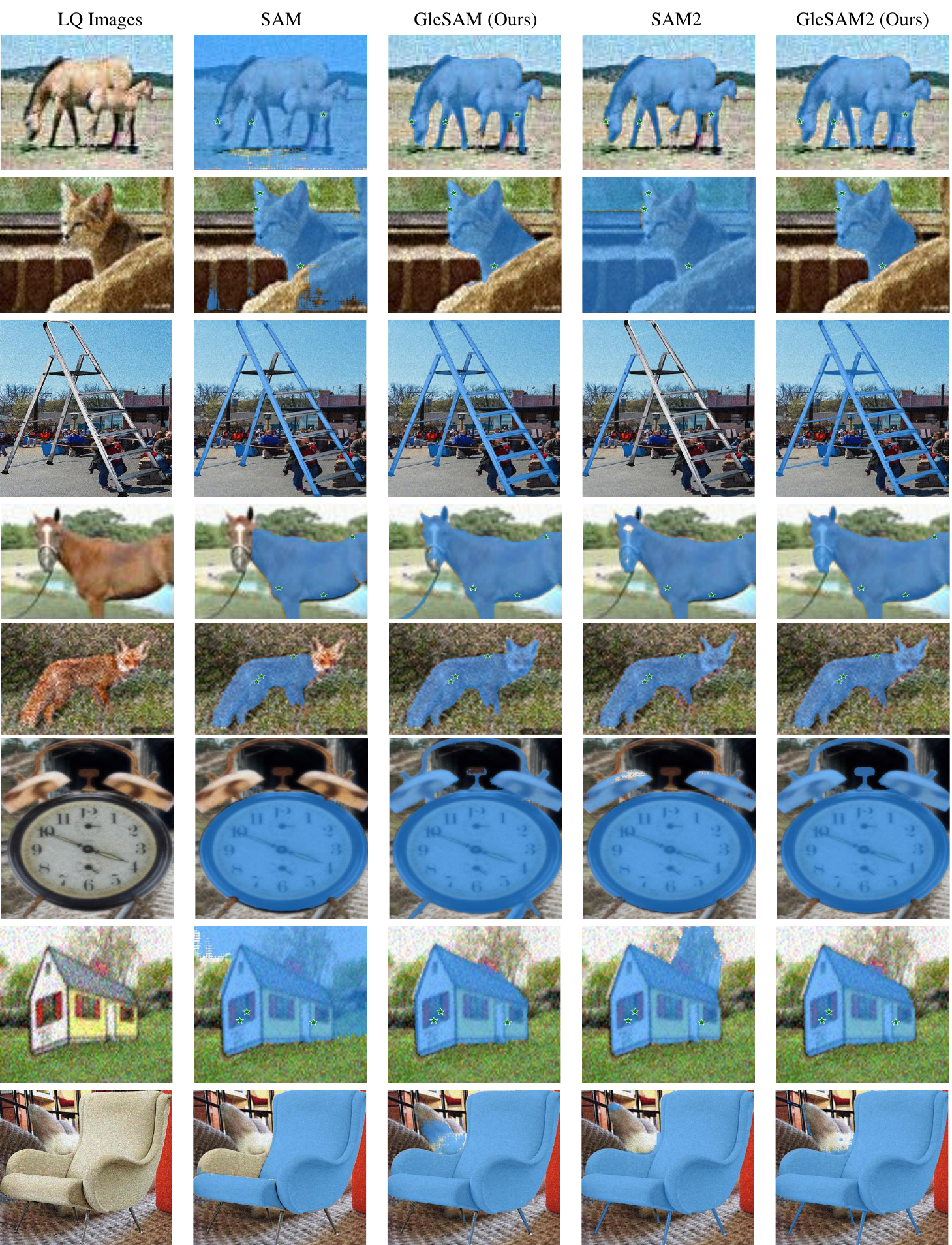}
   \caption{Qualitative Analysis of Segmentation: This figure offers a visual comparison to illustrate the enhanced performance of GleSAM and GleSAM2.}
   \label{fig:sm-vis}
\end{figure*}

\subsection{Comparison of Other Prompts}
In addition to point-based prompts, we conducted a comprehensive evaluation of our method using alternative prompting strategies, including GT-Box and Noise-Box. The GT-Box is directly derived from the ground truth mask and the Noise-Box introduces random perturbations to the GT-Box with a noise scale of 0.2 \cite{dndetr}, simulating scenarios with imperfect or noisy input. 
As shown in Tab. \ref{tab:sm-box-prompt}, our method, GleSAM and GleSAM2, consistently outperforms baseline models across all levels of image degradation. This robustness stems from the enhanced latent space representations, which mitigate noise-induced ambiguities during segmentation.
This highlights the adaptability and effectiveness of our method when dealing with varying prompt types.

\subsection{Comparison of Varying Number of Point Prompts} Fig. \ref{fig:sm-inter-seg} presents the interactive segmentation performance using point prompts. This comparison assesses our method and SAM with a range of input point numbers on the ECSSD dataset. 
GleSAM and GleSAM2 consistently outperform SAM and SAM2 across different numbers of point prompts (from 1 point to 10 points). Note that as the prompt contains less ambiguity (with more input points), the relative performance improvement becomes more significant. This indicates GleSAM's robust segmentation capability.

\subsection{Visualization of Feature Representation}
To evaluate the capability of our method in reconstructing high-quality representations from degraded inputs, we visualize the feature maps of low-quality images (LQ-Feat), the feature maps of original clear images (HQ-Feat), and the features reconstructed by our method (Ours-Feat). As shown in Fig. \ref{fig:sm-lqfeat}, the features extracted from low-quality images exhibit significant distortion and reduced clarity compared to their high-quality counterparts. 
In contrast, the features reconstructed by our method closely resemble the ``HQ-Feat'', effectively recovering structural and semantic details. This demonstrates the effectiveness of our approach in enhancing feature representations and ensuring robustness in degraded scenarios.

\subsection{Qualitative Results}
Fig. \ref{fig:sm-vis} presents a qualitative comparison of segmentation results produced by SAM, SAM2, and our proposed GleSAM and GleSAM2, on low-quality images. Due to the challenging degradations, SAM struggles to segment these objects accurately, resulting in serious detail missing and erroneous background prediction, showing its limitations. In contrast, GleSAM and GleSAM2 effectively recover finer details and achieve more precise segmentation results. 
These visual results underscore GleSAM's robustness in diverse real-world scenarios.

\section{Low-Quality Segmentation Dataset}
\label{sec:sm-lq-dataset}
Previous methods typically generate degraded images that contain only a single type of degradation.  However, \textbf{\textit{real-world degradation is often too complex to be accurately modeled using a single type of degradation.}} It frequently consists of a complex combination of various types of noise \cite{image-noise, image-noise2, image-noise3}. 
The lack of suitable data is one of the challenges we face in this task.  To address this issue, we seek to construct a comprehensive degraded image segmentation dataset that encompasses more complex degradations, rather than relying on a single type of degradation for each image. Motivated by previous methods \cite{robustsam, imagenet-c, syn-blur, Deblurring}, we opt to generate synthetic data to create a dataset for low-quality image segmentation.
In this section, we add details of degenerate modeling.

\begin{figure*}[!h]
  \centering
   \includegraphics[width=1\linewidth]{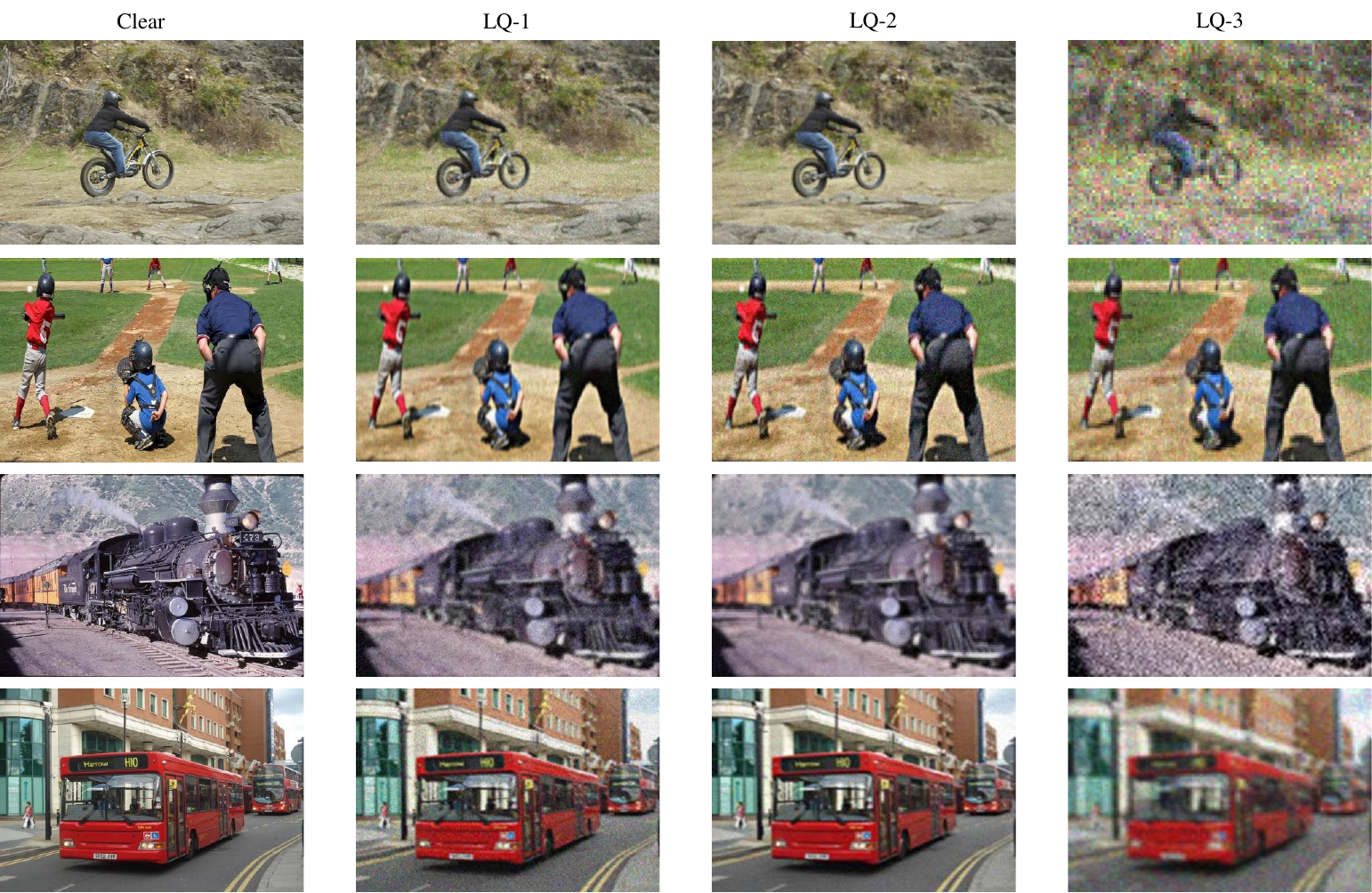}
   \caption{Examples from the LQ-Seg dataset illustrating images with varying levels of synthetic degradation: LQ-1, LQ-2, and LQ-3. These samples showcase the progressive quality deterioration used for evaluating the robustness of segmentation models.}
   \label{fig:sm-lqimg}
\end{figure*}

\subsection{{Degradation Modeling}}
Blur, downsampling, noise, and compression are the four key factors that contribute to the degradation of real images \cite{bsrgan}. Real image noise may consist of a complex combination of these degradations.  
To model a more practical degradation process, inspired by the previous work in image reconstruction \cite{realesrgan, bsrgan}, we model the degradation process $\mathcal{P}$ as the random combination of the above common degradations, including \textit{Blur} $\mathbf{B}$, \textit{Random Resize} ${\mathbf{R}}$, \textit{Noise} ${\mathbf{N}}$, and \textit{JPEG Compression} ${\mathbf{C}}$. It can be formulated as:
\begin{equation}
y=\mathcal{P}(x)=[\mathbf{B}, \mathbf{R}, \mathbf{N}, \mathbf{C}](x).
\end{equation}
Specifically, 1) For \textit{\textbf{Blur}} degradation, it is typically modeled as a convolution with a blur kernel. We randomly choose Gaussian kernels, generalized Gaussian kernels, and plateau-shaped kernels, with preset probability, kernel size, and standard deviation. 2) For \textit{\textbf{Random Resize}} operation, we consider both upsampling and downsampling operations with preset resize scales and randomly selected resize algorithms (\textit{i.e.}, bilinear interpolation, bicubic interpolation, and area resize). The randomness benefits include more diverse and complex resize effects. 3) For \textit{\textbf{Noise}} degradation, we consider two commonly used noise types: Gaussian noises and Poisson noises. Gaussian noise has a probability density function equal to that of the Gaussian distribution. The noise intensity is controlled by the standard deviation of the Gaussian distribution. Poisson noise follows the Poisson distribution, which is usually used to approximately model the sensor noise caused by statistical quantum fluctuations, that is, variation in the number of photos sensed at a given exposure level \cite{realesrgan}. 4) For \textit{\textbf{JPEG Compression}} operation, we use the off-the-shelf algorithms \cite{jpeg}, with a preset quality factor range. 

In addition, to enrich the granularity of degradation,  we employ multi-level degradation by adjusting the downsampling rates. We employed three different resize rates, \textit{i.e.}, [1, 2, 4], which correspond to three degradation levels from slight to severe: LQ-1, LQ-2, and LQ-3. Fig. \ref{fig:sm-lqimg} shows sample images with varying levels of synthetic degradation from the LQ-Seg dataset, demonstrating the diversity of degradation.

\subsection{Detailed Settings}
1) For \textbf{\textit{Blur}} process, the noise sigma range and Poisson noise scale are set to [1, 30] and [0.05, 3], respectively.
The blur kernel size is randomly selected from {7,9, ...21}. Blur standard deviation is sampled from [0.2, 3]. Shape parameter $\beta$ is sampled from [0.5, 4] and [1, 2] for generalized Gaussian and plateau-shaped kernels, respectively. We also use $sinc$ kernel with a probability of 0.1. We skip the second blur degradation with a probability of 0.2. 2) For \textbf{\textit{Noise}}, we adopt Gaussian noises and Poisson noises with a probability of $\{0.5, 0.5\}$. The noise sigma range and Poisson noise scale are set to [1, 30] and [0.05, 3], respectively. The gray noise probability is set to 0.4. 3) For \textbf{\textit{Random Resize}} operation, we randomly select a resize algorithm in area-resize, bilinear interpolation, and bicubic interpolation. 4) For \textbf{\textit{JPEG Compression}}, its quality factor is set to [30, 95]. More details can be found in the soon-released codes.

\section{More Implementation Details}
\label{sec:sm-detail}
\subsection{Training and Inference Details}
Built upon SAMs, GleSAM/GleSAM2 inherits prompt-based segmentation. During training, we utilize random points or the bounding box as prompts, which are encoded into prompt vectors by the frozen prompt encoder and then fed into the decoder.
Our model is trained using the AdamW optimizer with the learning rate of $1\times10^{-4}$ and batch size of 4. 
The pre-trained U-Net in Stable Diffusion (SD) 2.1-base \cite{ldm} is adopted as the denoising backbone. We set the $T=1000$ in Eq. 3 as default.
Our approach can be efficiently trained on 4$\times$ A100 GPUs within approximately 30 hours, during which we fine-tune the U-Net for 100K iterations and the decoder for only 20K iterations.          
During the inference, our methods follow the same pipeline as SAM, ensuring compatibility and ease of deployment. Detailed training and inference schemes are shown in Algorithm \ref{al:train} and \ref{al:inference}.

\subsection{Prompt Generation}
For box-based evaluation, we use the ground truth mask to generate the bounding box and input it as the box prompt. For noise-box-based evaluation, the noise-box is generated by adding noise to the GT box as the prompt input, following \cite{dndetr}.
In our experiments, the noise scale is set to 0.2 by default. For point-based evaluation, we randomly sample several points from the ground truth masks and use them as the input prompt. In our experiments, the number of random points is set to 3 by default.





\begin{center}
\begin{algorithm}
\caption{\textbf{Training Scheme of GleSAMs}}
\label{al:train}
\begin{algorithmic}[1]
\State \textbf{Input:} Training dataset $\mathcal{S}$, pretrained SAM including image encoder $\mathcal{E}_{\theta}$, prompt encoder $\mathcal{P}_{\theta}$, and mask decoder $\mathcal{D}_{\theta}$. Pretrained U-Net $\epsilon_{\theta}$ with learnable LoRA layers, fine-tuning iteration $N_1, N_2$. \\
\textit{/*               Fine-tuning U-Net                */}
\For{$i \leftarrow 1$ to $N_1$}
    \State Sample $x_H, x_L$ from $\mathcal{S}$
    \State \textit{/* Network forward */}
    \State $[z_H, z_L] \leftarrow \mathcal{E}_{\theta}([x_H, x_L])$
    \State $\hat{z}_H \leftarrow \mathrm{GLE}(z_L)$
    \State \textit{/* Compute reconstructive loss */}
    \State $\mathcal{L}_{\mathrm{Rec}} = \mathcal{L}_{\text{MSE}}(\hat{z}_H, z_H)$
    \State \textit{/* Network parameter update */}
    \State Update learnable parameters with $\mathcal{L}_{\mathrm{Rec}}$
\EndFor \\
\textit{/* Fine-tuning Decoder */}
\For{$i \leftarrow 1$ to $N_2$}
    \State Sample $\hat{z}_{H}, m_g$ from $\mathcal{S}$
    \State \textit{/* Network forward */}
    \State Sample prompts $p$ from $m_g$ 
    \State $m_p \leftarrow \mathcal{D}_{\theta}(\hat{z}_{H}, \mathcal{P}_{\theta}(p))$
    \State \textit{/* Compute segmentation loss */}
    \State $\mathcal{L}_{\mathrm{Seg}} = \mathcal{L}_{\text{Dice}}(m_p, m_g) + \mathcal{L}_{\mathrm{Focal}}(m_p, m_g)$
    \State \textit{/* Network parameter update */}
    \State Update learnable parameters with $\mathcal{L}_{\mathrm{Seg}}$
\EndFor
\State \textbf{Output:} Fine-tuned U-Net $\epsilon_{\theta}$ and mask decoder $\mathcal{D}_{\theta}$
\end{algorithmic}
\end{algorithm}
\end{center}

\begin{algorithm}
\caption{\textbf{Inference Scheme of GleSAMs}}
\label{al:inference}
\begin{algorithmic}[1]
\State \textbf{Input:} Low-quality image $x_L$ and corresponding prompt $p$, Pretrained image encoder $\mathcal{E}_{\theta}$ and prompt encoder $\mathcal{P}_{\theta}$. Fine-tuned mask decoder $\mathcal{D}_{\theta}$ and U-Net $\epsilon_{\theta}$.
\State \textit{/* Image encoding */}
\State $z_L \leftarrow \mathcal{E}_{\theta}(x_L)$
\State \textit{/* Generative latent space enhancement */}
\State $\hat{z}_H \leftarrow \mathrm{GLE}(z_L)$
\State \textit{/* Mask decoding */}
\State $m_p \leftarrow D_{\theta}(\hat{z}_{H}, \mathcal{P}_{\theta}(p))$
\State \textbf{Output:} Predicted mask $m_p$.
\end{algorithmic}
\end{algorithm}



\section{More qualitative results.}
\label{sec: sm-vis}

\subsection{Visual Results on Other Degradations}
\begin{figure*}[!h]
  \centering
\includegraphics[width=0.75\linewidth]{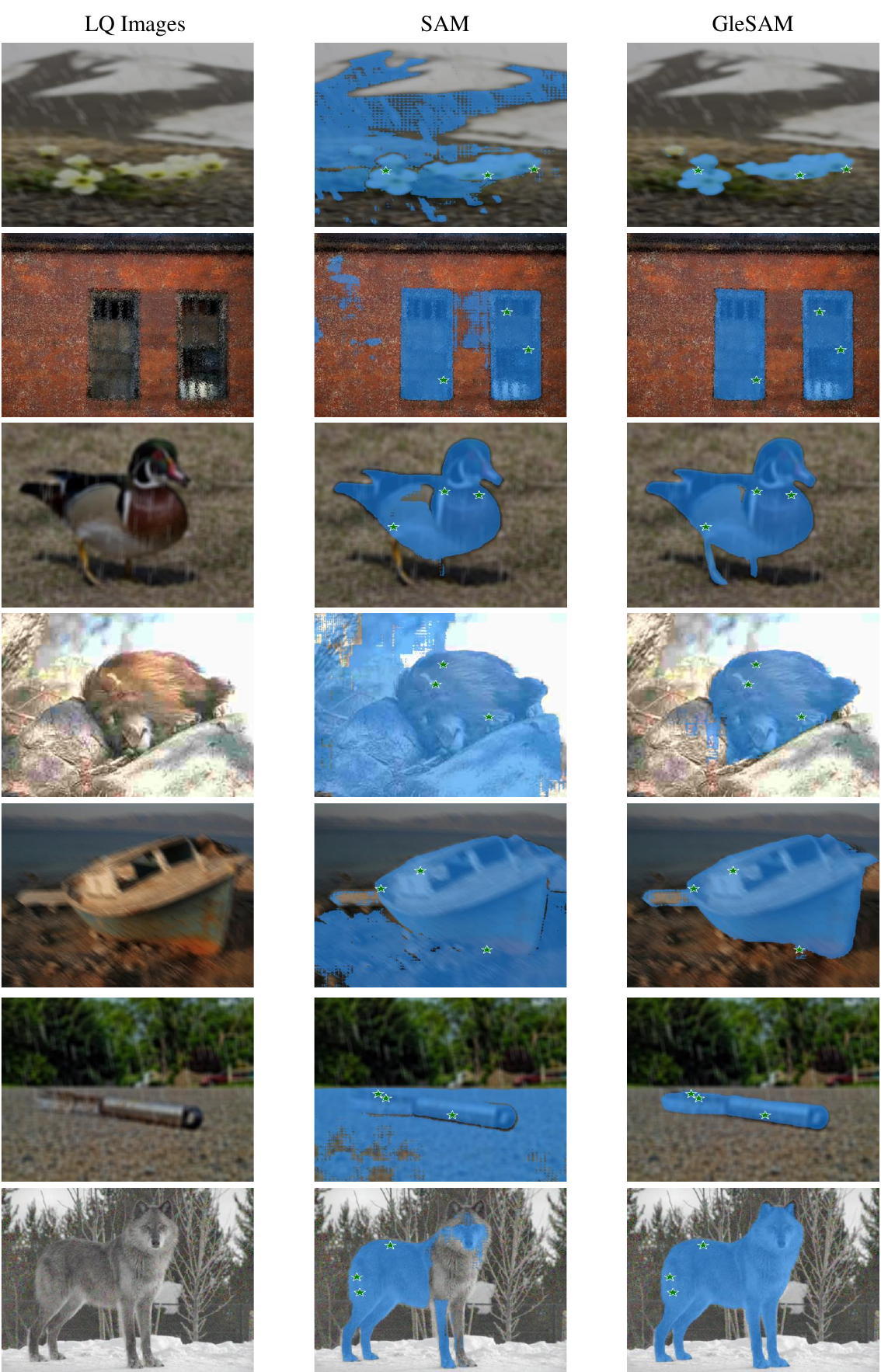}
   \caption{Visual comparisons of SAM and GleSAM on the unseen ECSSD dataset under RobustSeg-style degradations, such as rain, snow, low-light conditions, and others. The results demonstrate the superior generalization capability of GleSAM to handle unseen degradations not included in the training set.}
   \label{fig:sm-sam-robseg-vis}
\end{figure*}

\begin{figure*}[!h]
  \centering
\includegraphics[width=0.75\linewidth]{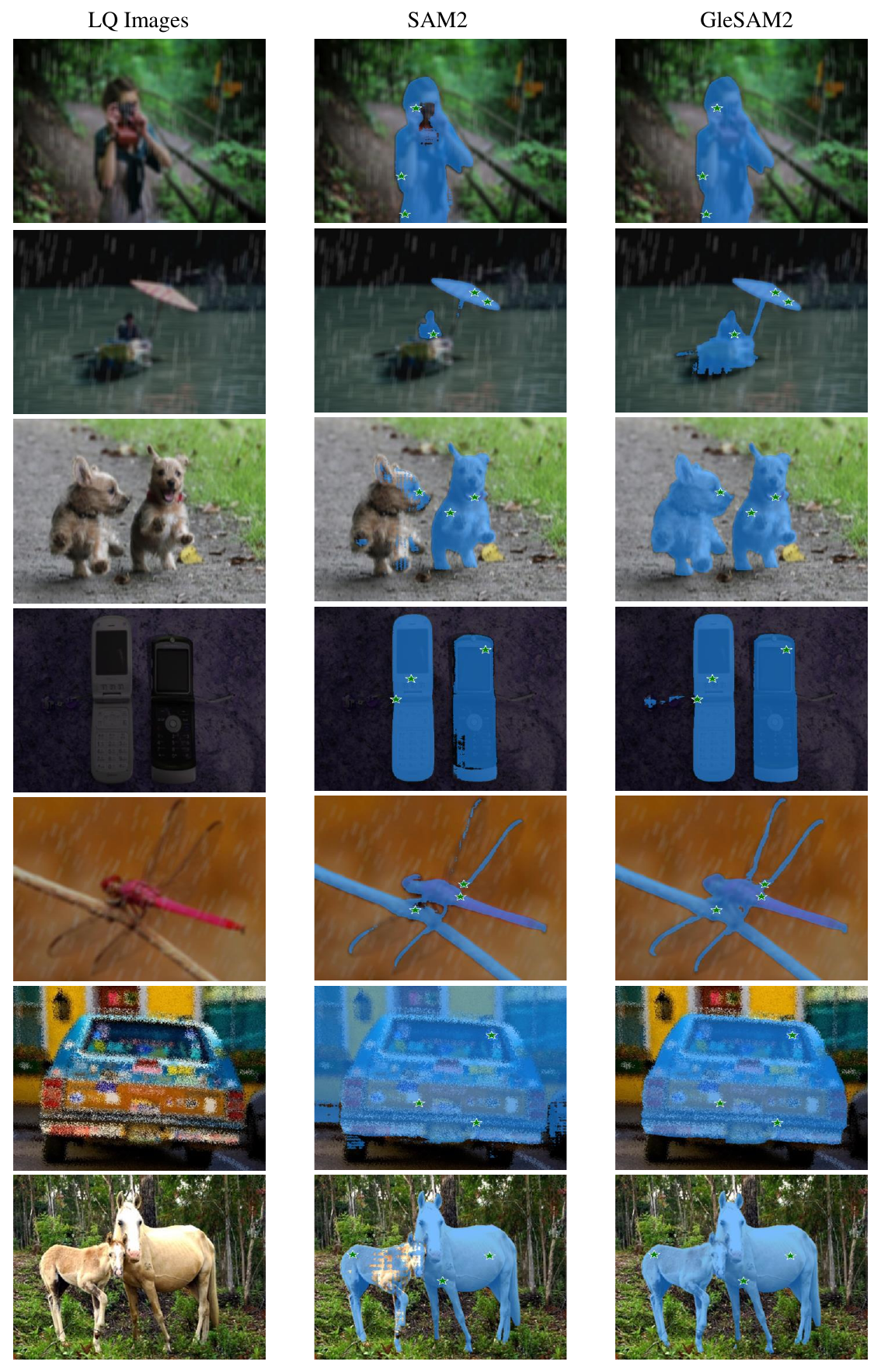}
   \caption{Visual comparisons of SAM2 and GleSAM2 on the unseen ECSSD dataset under RobustSeg-style degradations, such as rain, snow, low-light conditions, and others. The results demonstrate the superior generalization capability of GleSAM2 to handle unseen degradations not included in the training set.}
   \label{fig:sm-sam2-robseg-vis}
\end{figure*}

Fig. \ref{fig:sm-sam-robseg-vis} and Fig. \ref{fig:sm-sam2-robseg-vis} highlight the segmentation performance of SAM/SAM2 and GleSAM/GleSAM2 on the unseen ECSSD dataset under various RobustSeg-style degradations, including rain, snow, low-light. 
These degradation types were not encountered during training, providing a robust evaluation of the model's generalization. 
In contrast, our methods consistently produce high-quality segmentation masks in various degraded environments. These results underscore GleSAM's adaptability and robustness when dealing with unseen degradation types, making it highly suitable for real-world applications with diverse image qualities.

\subsection{More Visual Comparisons on Unseen Dataset.}
\begin{figure*}[!h]
  \centering
\includegraphics[width=0.8\linewidth]{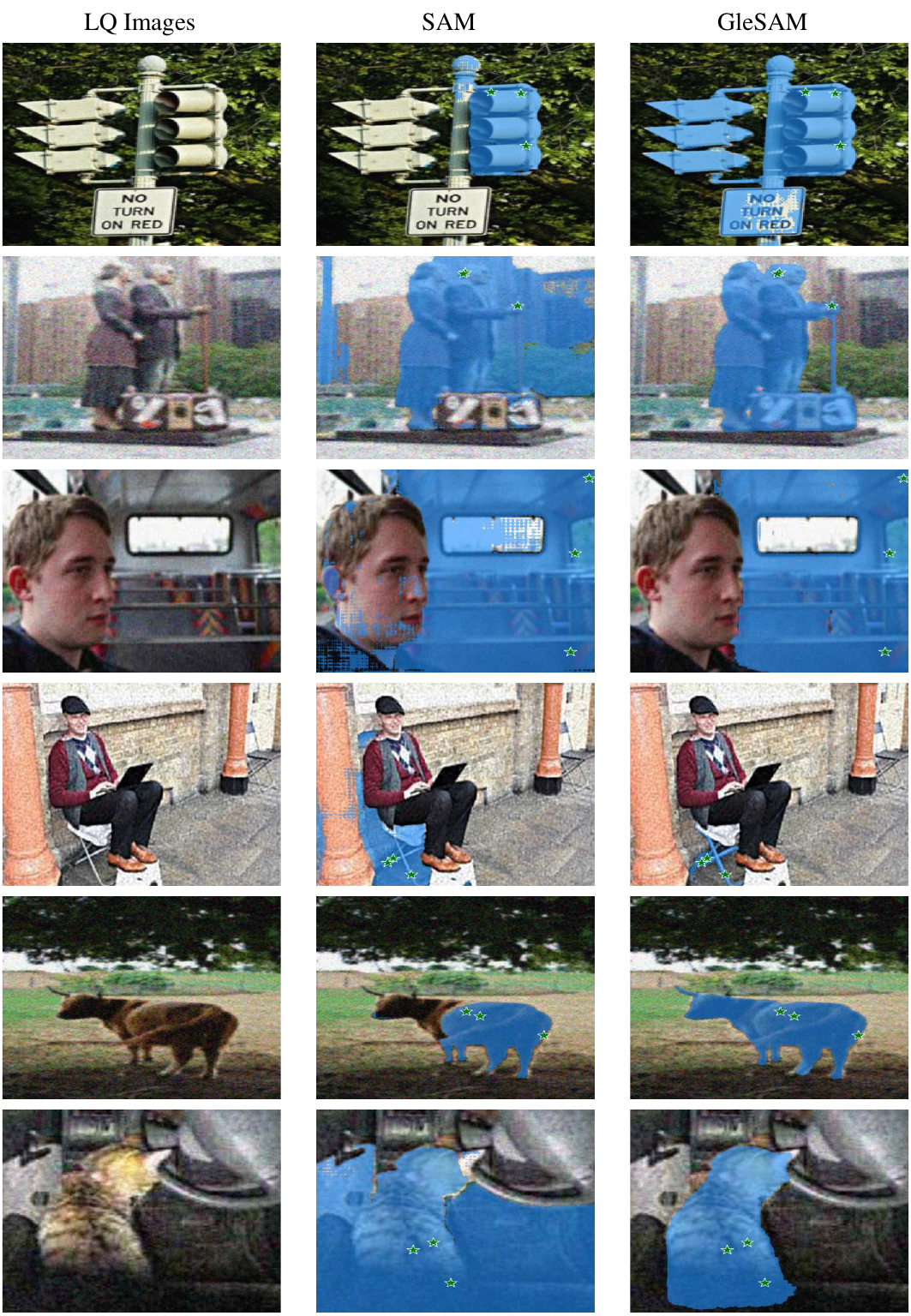}
   \caption{Visual comparisons of segmentation results on unseen COCO dataset. This figure illustrates the enhanced performance of GleSAM.}
   \label{fig:sm-sam-coco-vis}
\end{figure*}

\begin{figure*}[!h]
  \centering
\includegraphics[width=0.8\linewidth]{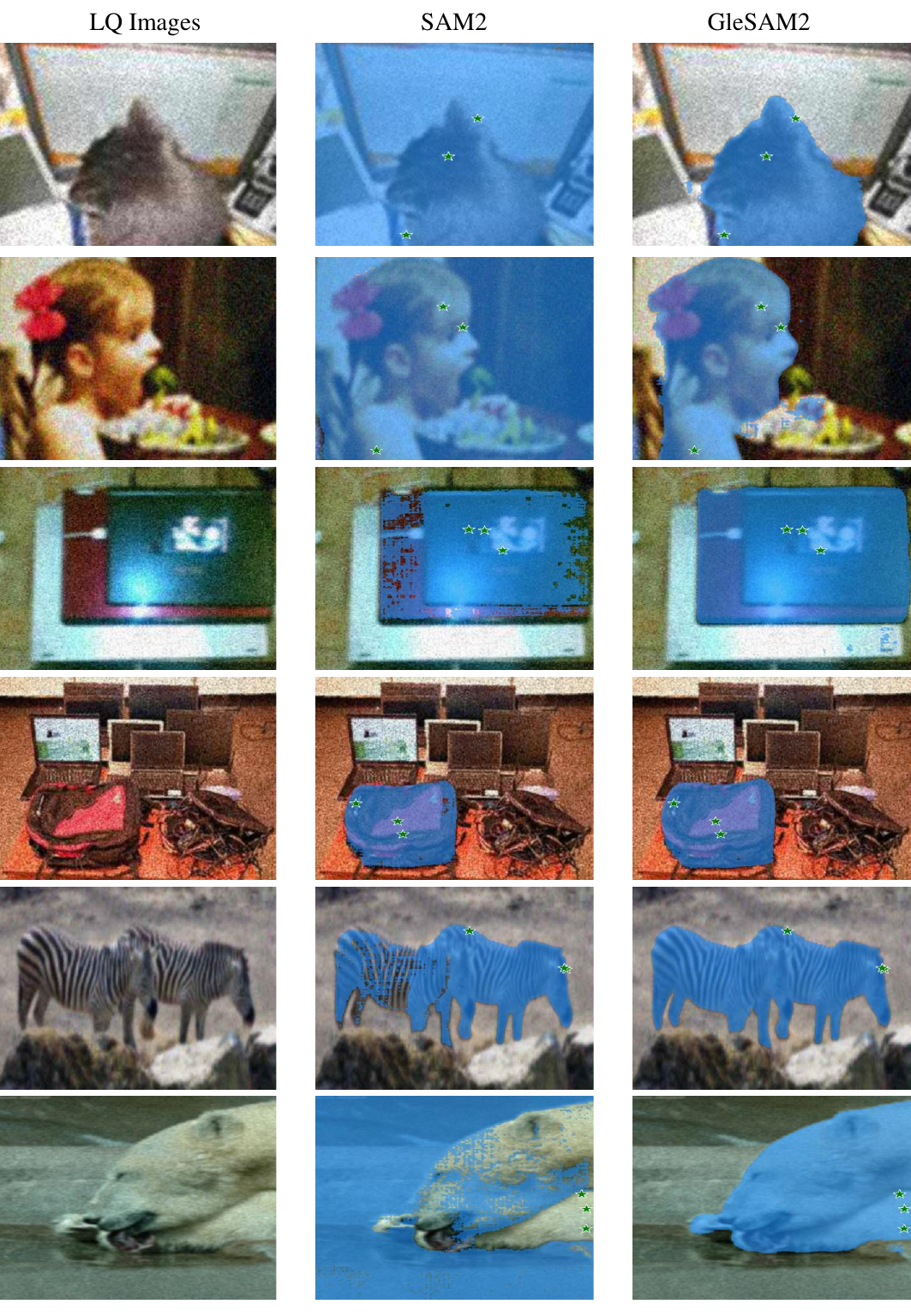}
   \caption{Visual comparisons of segmentation results on unseen COCO dataset. This figure illustrates the enhanced performance of GleSAM2.}
   \label{fig:sm-sam2-coco-vis}
\end{figure*}

Fig. \ref{fig:sm-sam-coco-vis} and Fig. \ref{fig:sm-sam2-coco-vis} showcase the segmentation performance of SAM/SAM2 and our proposed GleSAM/GleSAM2 on the unseen COCO \cite{coco} dataset. Each row presents examples of low-quality images, followed by the corresponding segmentation outputs. SAM and SAM2 struggle to generate accurate masks under severe degradations, often failing to capture object boundaries and details. In contrast, our GleSAM and GleSAM2 produce more precise segmentation masks, highlighting the robustness of our method in processing low-quality images.

\section*{Acknowledgements}
This work was supported in part by the National Natural
Science Foundation of China (NSFC) under Grant 62372382.

{
    \small
    \bibliographystyle{ieeenat_fullname}
    \bibliography{main}
}

\end{document}